\documentclass[sigconf]{acmart}
\usepackage[utf8]{inputenc}
\usepackage{newunicodechar}
\usepackage{enumitem}
\usepackage[table]{xcolor}
\usepackage{multirow}
\usepackage{subcaption}
\usepackage{arydshln}
\usepackage{comment}
\newunicodechar{：}{:}
\newcolumntype{L}[1]{>{\raggedright\arraybackslash}p{#1}}

\usepackage[linesnumbered,ruled,vlined]{algorithm2e}
\usepackage{amsmath}

\SetKwInput{KwParam}{Hyperparameters}

\AtBeginDocument{%
  }

\setcopyright{none}
\copyrightyear{2026}
\acmYear{2026}
\acmDOI{}
\acmConference[KDD '26]{ACM SIGKDD}{August, 2026}{Jeju, Korea}
\acmISBN{}

\begin{document}

\title{U-CAN: Utility-Aware Contrastive Attenuation for Efficient Unlearning in Generative Recommendation}

\author{Zezheng Wu}
\email{wzz@mails.guet.edu.cn}
\affiliation{
  \institution{Guilin University of Electronic Technology}
  \city{Guilin}
  \country{China}}

\author{Rui Wang}
\email{wangrui377@outlook.com}
\affiliation{
  \institution{University of Southampton}
  \city{Southampton}
  \country{United Kingdom}}

\author{Xinghe Cheng}
\email{jnuchengxh@hotmail.com}
\affiliation{
  \institution{Jinan University}
  \city{Guangzhou}
  \country{China}}

\author{Yang Shao}
\email{Syyyang@mails.guet.edu.cn}
\affiliation{%
  \institution{Guilin University of Electronic Technology}
  \city{Guilin}
  \country{China}}

\author{Qing Yang}
\email{gtyqing@hotmail.com}
\affiliation{%
  \institution{Guilin University of Electronic Technology}
  \city{Guilin}
  \country{China}}

\author{Jiapu Wang}
\email{jiapuwang9@gmail.com}
\authornotemark[1]
\affiliation{%
  \institution{Nanjing University of Science and Technology}
  \city{Nanjing}
  \country{China}}

\author{Jingwei Zhang}
\email{gtzjw@hotmail.com}
\authornote{Corresponding author.}
\affiliation{%
  \institution{Guilin University of Electronic Technology}
  \city{Guilin}
  \country{China}}



\begin{abstract}
Generative Recommendation (GenRec) typically leverages Large Language Models (LLMs) to redefine personalization as an instruction-driven sequence generation task. 
However, fine-tuning on user logs inadvertently encodes sensitive attributes into model parameters, raising critical privacy concerns. Existing Machine Unlearning (MU) techniques struggle to navigate this tension due to the \textit{Polysemy Dilemma}, where neurons superimpose sensitive data with general reasoning patterns, leading to catastrophic utility loss under traditional gradient or pruning methods. 
To address this, we propose \textbf{\underline{U}}tility-aware \textbf{\underline{C}}ontrastive \textbf{\underline{A}}ttenuatio\textbf{\underline{N}} (U-CAN), a precision unlearning framework that operates on low-rank adapters. 
U-CAN quantifies risk by contrasting activations and focuses on neurons with asymmetric responses that are highly sensitive to the forgetting set but suppressed on the retention set. To safeguard performance, we introduce a utility-aware calibration mechanism that combines weight magnitudes with retention-set activation norms, assigning higher utility scores to dimensions that contribute strongly to retention performance.
Unlike binary pruning, which often fragments network structure, U-CAN develop adaptive soft attenuation with a differentiable decay function to selectively down-scale high-risk parameters on LoRA adapters, suppressing sensitive retrieval pathways and preserving the topological connectivity of reasoning circuits.
Experiments on two public datasets across seven metrics demonstrate that U-CAN achieves strong privacy forgetting, utility retention, and computational efficiency\footnote{\href{https://anonymous.4open.science/r/U-CAN-7D6F}{Code is available: https://anonymous.4open.science}}.

\end{abstract}


    



\maketitle

\section{Introduction}
Generative Recommendation (GenRec)~\cite{ZhuYW2025, YueRM2023, GengLF2022, KeDL2024} is a paradigm shift in recommender systems, where Large Language Models (LLMs) serve as the generative backbone to model user intent and item semantics, enabling recommendations via semantic reasoning rather than pure ranking~\cite{pan2024unifying,zhao2023survey,naveed2025comprehensive}.
When models are fine tuned on user interactions, improved personalization arises from parameters encoding user specific preferences and attributes, which in turn raises the risk of attribute inference or data extraction, and the central challenge is to remove user dependent information while preserving general recommendation capability~\cite{CarliniTW2021, NguyenHR2025, ANGZW2025, WaHZ2025, NasrCH2023}.


\begin{figure}[t]
  \centering
  \includegraphics[width=0.47\textwidth]{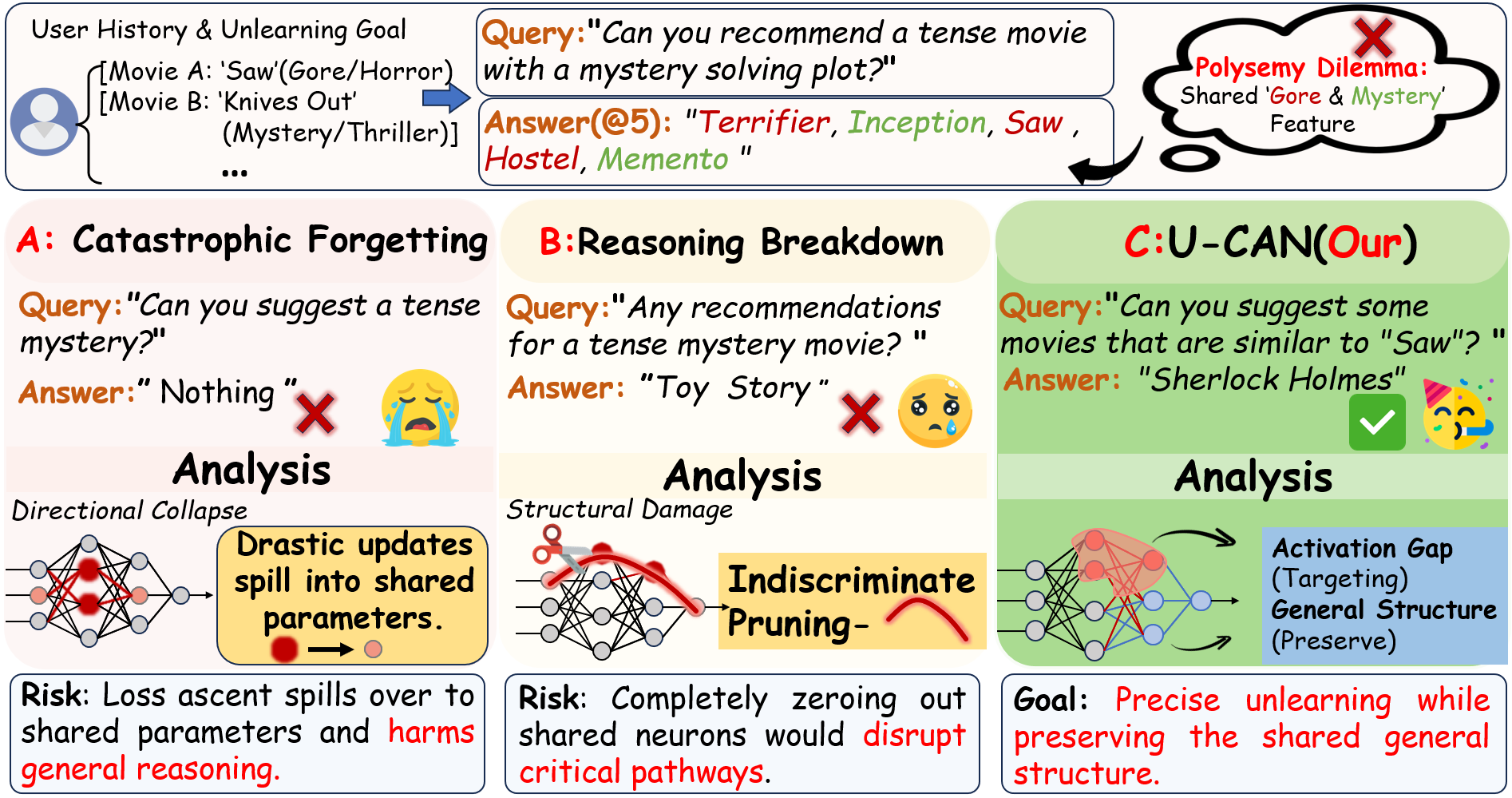}
  \Description{The \textit{Polysemy Dilemma} in unlearning.}
  \caption{Comparison of our proposed \texttt{U-CAN} (C) with traditional methods (A,B). The traditional methods either distort shared parameters via gradient updates or break functional pathways via hard pruning, whereas we locate risky neurons via activation difference comparison and suppress high-risk parameters with continuous soft decay, achieving more precise forgetting while preserving general reasoning ability.
}
\vspace{-20pt}
  \label{fig:motivation}
\end{figure}

Machine Unlearning (MU)~\cite{BourtouleCC2021, JangYY2023, WangZG2025, TaoWY2025} addresses this need by removing the influence of designated training data from a trained model, aiming for targeted forgetting without sacrificing model utility.
However, existing methodologies~\cite{MainiFS2024, DaiDH2022, MaFW2023, ZhangZZ2025} frequently encounter a fundamental tension between forget and utility. 
As depicted in Fig.~1(A), gradient based approaches~\cite{ChCH2025, CSC2022}, despite their adaptability, often suffer from ``\textit{Directional Collapse}'', while their adaptive parameter updates spill into shared reasoning representations and lead to a sharp drop in recommendation quality.
Meanwhile, as depicted in Fig.~1(B), pruning-based strategies~\cite{ZhangZZ2025}, despite being effective at identifying high saliency parameters, often suffer from ``\textit{Structural Damage}'', while their indiscriminate zeroing out fragments the model structure and severs the functional pathways essential for general reasoning.

Crucially, in GenRec models, parameters rarely sequester privacy information in isolation. Instead, sensitive concepts are encoded within superposed activations that concurrently facilitate general linguistic syntax, narrative coherence, and domain-specific knowledge. This intrinsic entanglement constitutes a \textit{Polysemy Dilemma}: parameters that respond strongly to privacy-related interactions concurrently determine the quality of ordinary recommendations. As elucidated by the \textit{Catastrophic Forgetting} panel in Figure~\ref{fig:motivation}, sensitive features exist as distributed superpositions across syntactic neurons rather than distinct modular units \cite{GevaSB2021}. Within this highly entangled space, gradient-ascent methods \cite{CSC2022, ZhLB2024} attempt to reverse the influence of the forget set, yet these updates navigate exceedingly sharp trajectories within the coupled loss landscape, causing perturbations to propagate from privacy-critical regions into shared functional reasoning spaces.

Pruning-based methods instead identify high-saliency units or weights for binary deletion. As illustrated by the \textit{Reasoning Breakdown} panel in Figure~\ref{fig:motivation}, such hard deletion breaks activation pathways at the structural level. Under the polysemy regime, the pruned neurons also carry non-sensitive semantics \cite{FrankleC2019, WeiWS2022}, so indiscriminate zeroing reduces utility even when privacy leakage is mitigated. Both classes of methods therefore lack a mechanism that can localize risk at the level of fine-grained activations while preserving the topological connectivity that supports overall performance.

To bridge the gap between precise erasure and structural stability, we propose \textbf{\underline{U}}tility-aware \textbf{\underline{C}}ontrastive \textbf{\underline{A}}ttenuatio\textbf{\underline{N}} (U-CAN), an unlearning framework tailored for generative recommendation.
Unlike approaches that directly threshold raw activations or rely on rigid binary masks, U-CAN combines contrastive activation analysis with structure-preserving parameter approximation.
We introduce a contrastive activation-difference score to quantify how strongly each neuron responds to sensitive versus general inputs, which enables precise isolation of risk-related neurons.
To reconcile privacy erasure and utility maintenance, a utility-aware calibration module integrates weight magnitudes with activation norms to protect parameters that are important for reasoning.
Finally, instead of hard pruning, U-CAN applies an adaptive soft attenuation scheme that uses a continuous decay function to down-scale high-risk parameters on LoRA adapters.
This design supports one-shot suppression of privacy-related responses while preserving the connectivity of reasoning pathways that sustain general utility.

The main contributions of this paper are summarized as follows:
\begin{itemize}[left=0cm]
\item We introduce U-CAN, an unlearning framework for GenRec that feature a synergistic dual-screening mechanism. By harmonizing contrastive activation analysis with utility-aware structural calibration, U-CAN effectively disentangles privacy-sensitive responses from essential reasoning behaviors within the entangled representation space;
\item We develop an adaptive soft attenuation strategy based on a differentiable decay function. Unlike rigid binary pruning, this mechanism facilitates precise down-scaling of high-risk parameters on LoRA adapters, successfully suppressing sensitive retrieval pathways while rigorously preserving the topological connectivity of underlying reasoning circuits in the model;
\item We provide a comprehensive empirical validation across diverse public datasets. Our results demonstrate that U-CAN achieves excellent unlearning performance across seven critical evaluation metrics, attaining a strong balance of privacy forgetting, utility retention, and one-shot operational efficiency without requiring secondary training.

\end{itemize}

\begin{figure*}[tbp]
  \centering
  \includegraphics[width=1.0\textwidth]{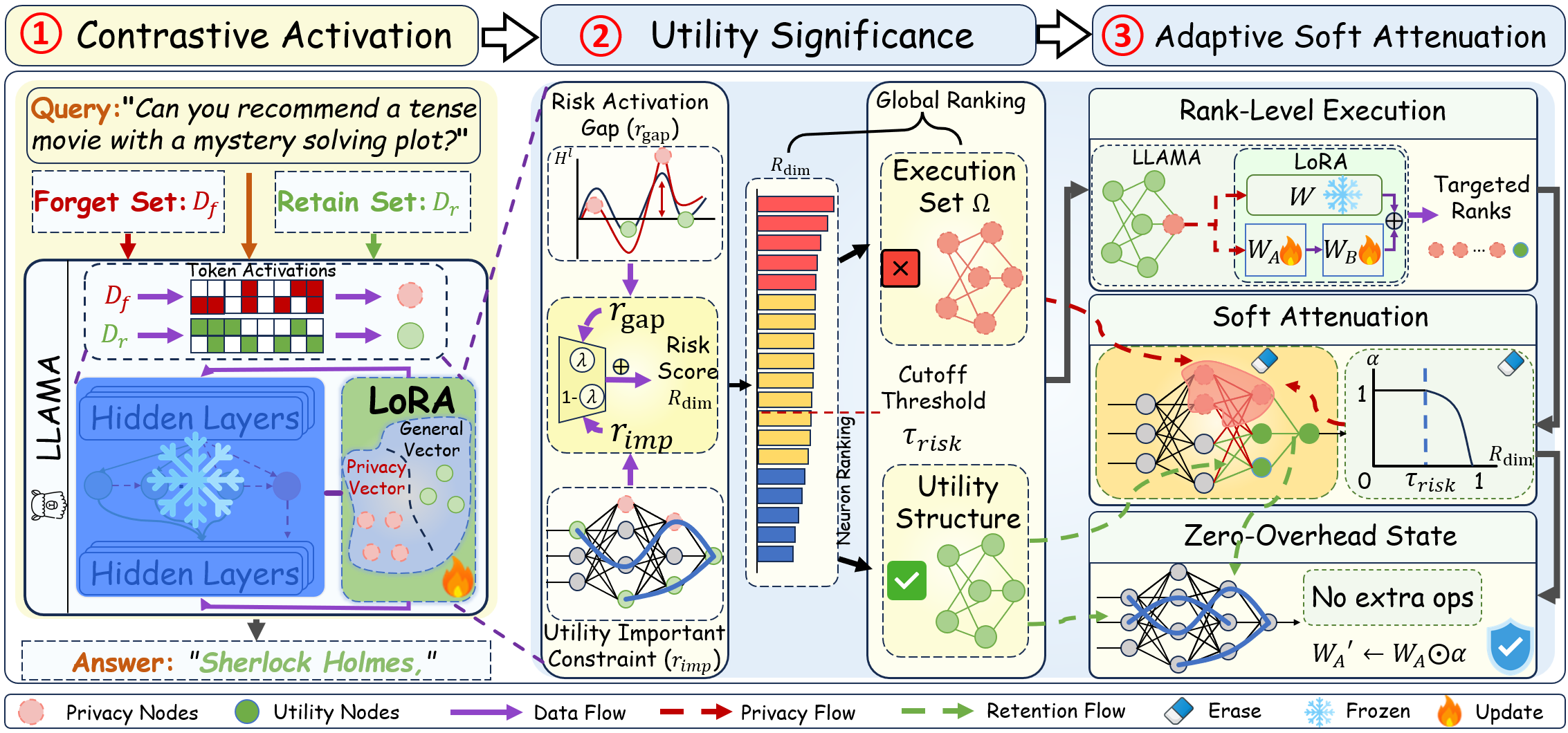}
  \Description{This is a flowchart of the overall framework.}
  \caption{The overall framework of U-CAN. The pipeline orchestrates three integral modules. (1) Contrastive Activation isolates entangled features by leveraging activation gaps to pinpoint privacy-sensitive neurons. (2) Utility Significance quantifies parameter importance by fusing static weight magnitudes with dynamic input intensities, ensuring core capabilities remain intact. (3) Adaptive Soft Attenuation scales down risk parameters using a continuous decay curve, maintaining network connectivity and avoiding the abrupt damage caused by binary pruning.}
  \label{fig:framework}
\end{figure*}
\section{Related Work}
Unlearning in GenRec can be discussed along two complementary lines: one reviews the current landscape of GenRec and its privacy-related risk factors, 
and the other line investigates which machine unlearning approaches can effectively eliminate the impact of these risk factors.
Accordingly, this section focuses on two themes: (i) Generative Recommendation and (ii) Machine Unlearning.

\subsection{Generative Recommendation}
Generative Recommendation (GenRec) formulates recommendations as a language modeling task~\cite{ChCH2025,BaZZ2023,GuLY2024,ZhZW2025,YaLW2025,wang2023survey, wang2024large, wang2024ime,cheng2025education,cheng2025nr4der}: user interaction histories and item attributes are serialized into token sequences, and predictions are produced via autoregressive generation rather than through conventional embedding-based retrieval-and-ranking pipelines.
Within this framework, LLMs naturally underpin GenRec by reducing reliance on massive item-id embedding tables through tokenized item representations, thus improving coverage of long-tail and newly introduced items in certain settings.

Early representative work such as P5~\cite{GeLF2022} directly converts user interaction histories into natural language prompts, enabling a unified generative recommendation framework. 
Subsequent methods, including TallRec~\cite{BaZZ2023} and LLaRA~\cite{LiLY2024}, further incorporate parameter-efficient fine-tuning mechanisms such as Adapters or LoRA, which significantly reduce domain adaptation cost while preserving large model capability. Meanwhile, LlamaRec~\cite{YueRM2023} introduces an LLM-based two-stage generative recommendation framework, where decoupled candidate generation and reranking deliver strong performance and inference efficiency.

Although prior studies demonstrate the effectiveness of generative paradigms for recommendation, they primarily focus on modeling capability and training efficiency, leaving privacy risks underexplored, especially those arising when personalized fine-tuning memorizes sensitive user interactions~\cite{HuZX2025, NguyenHR2025}. Therefore, selectively removing the influence of privacy-related interactions from model behaviour while maintaining generative recommendation utility remains a central open challenge in GenRec.

\subsection{Machine Unlearning}
Machine Unlearning (MU) was originally proposed to comply with data protection regulations such as the \textit{right to be forgotten}~\cite{CaoY2015, BourtouleCC2021}. 
Its primary objective is to eliminate the influence of targeted training data points from the model parameters~\cite{NguyenHR2025, ZhangHB2024, DeHP2023, LiWY2024, 0012WYY0Y25, 0012YP24, ZhangWYPTP24}.
While foundational unlearning paradigms were predominantly tailored for discriminative tasks such as image classification~\cite{GinartGV2019}, the exponential proliferation of LLMs has necessitated a strategic pivot toward generative unlearning. 
This evolution addresses the multifaceted risks inherent in modern foundation models, ranging from the propagation of toxic content to severe privacy leakage and copyright violations~\cite{JangYY2023, CarliniTW2021, EldanR2024}.

The most fundamental paradigm in unlearning is Gradient Ascent (GA), which updates parameters by maximizing prediction loss on a designated \textit{forget set}~\cite{JangYY2023}.
However, GA is highly sensitive to hyperparameters and often triggers catastrophic forgetting on desired knowledge~\cite{FanLZ2023}. To stabilize this process, research has transitioned toward preference-based objectives~\cite{LuWH2022}. 
Specifically, Negative Preference Optimization (NPO) adapts the Direct Preference Optimization framework~\cite{RafailovSM2023} for unlearning. By anchoring optimization to a reference model, it effectively steers distributions away from undesirable outputs while precluding the instability associated with unbounded loss.
Other common baselines employ random labeling or mismatch objectives to replace specific knowledge with generic responses~\cite{LiuYJ2025}.

Recent studies emphasize that effective unlearning requires a precise understanding of the interaction between data and model components~\cite{LiuYJ2025, LiCZ2024}. 
This localization-based unlearning paradigm proceeds by identifying specific parameters or computational units important to the target knowledge~\cite{WuLX2023, FanLZ2023}. Building on this, Zhang et al.~\cite{ZhangZZ2025} introduced LLM-Eraser, a two-stage method that uses selective pruning and contrastive distillation to disentangle desired and undesired knowledge. 
By pruning only those parameters crucial to the undesired domain, such methods effectively minimize the performance deficit on general tasks.

The evaluation of LLM unlearning remains intrinsically challenging, as the definitive benchmark of exact retraining is computationally prohibitive for large-scale models~\cite{MainiFS2024}.
While conventional metrics quantify erasure within the target scope and utility on unrelated tasks, they frequently overlook the global parameter coupling induced by gradient updates~\cite{ZhangZZ2025}.
This gap is further amplified in Generative Recommendation, where personalized fine-tuning on sequential user interactions is performed under frequent updates and tight utility constraints, yet most existing studies emphasize modeling capacity and efficiency rather than how interaction traces are memorized, re-triggered, and reflected in recommendation-specific behaviors.
In particular, evaluation protocols for unlearning in the GenRec setting remain limited, where forgetting should be assessed not only by task-agnostic probes but also by changes in recommendation outputs attributable to the targeted interaction history.
Such an entanglement in high dimensions makes the isolation of sensitive features from the shared semantic space exceptionally difficult.
Consequently, indiscriminate \textit{hard pruning} strategies~\cite{MaFW2023} risk severing the dense synaptic pathways essential for general reasoning~\cite{ZhongWM2023}, precipitating significant structural degradation.

\section{Preliminaries and Problem Definition}
\textbf{Machine Unlearning (MU).} MU aims to surgically eliminate the influence of specific data from a pre-trained model $f_{\theta}$ without the need for full retraining. 
We define the original training set as $\mathcal{D} = \mathcal{D}_{r} \cup \mathcal{D}_{f}$, where $\mathcal{D}_{f}$ denotes the forgetting target and $\mathcal{D}_{r}$ represents the retention data. 
The core objective is to efficiently optimize the model parameters from $\theta$ to $\theta^{*}$ such that the information from $\mathcal{D}_{f}$ is effectively erased. 

Theoretically, the unlearning process is successful if the distribution of the updated parameters $\theta^{*}$ closely approximates that of a model retrained exclusively on the retention set. We formalize the retrain-on-retain reference as: 
\begin{equation}
\theta_{r} \;=\; \arg\min_{\theta}\; \mathcal{L}(\theta;\mathcal{D}_{r}),
\label{eq:retrain_on_retain}
\end{equation}
and require $P(\theta^{*}) \approx P(\theta \mid \mathcal{D}_{r})$. 
This distributional equivalence ensures erasure for privacy information associated with $\mathcal{D}_{f}$ and sustains generalization and utility within the retention set $\mathcal{D}_{r}$.

\textbf{Low-Rank Adaptation (LoRA).}
LoRA~\cite{HuSW2022} is a parameter-efficient adaptation mechanism on top of a frozen pre-trained backbone.
Let the backbone parameters be $\theta_{0}$ and the trainable adapter parameters be $\phi$.
Consider a representative linear layer with input $x$ and output $o$.
LoRA keeps the original weight matrix $W_{0}\in\mathbb{R}^{d_{1}\times d_{2}}$ fixed and adds a rank-$r$ update parameterized by two learnable matrices $W_{B}\in\mathbb{R}^{d_{1}\times r}$ and $W_{A}\in\mathbb{R}^{r\times d_{2}}$:
\begin{equation}
o \;=\; W_{0}x \;+\; W_{B}W_{A}x,
\label{eq:lora}
\end{equation}
where $r\ll \min(d_{1},d_{2})$.
This design introduces $d_{1}r+rd_{2}$ trainable parameters, which is substantially smaller than the $d_{1}d_{2}$ parameters of the full matrix $W_{0}$.
During adaptation, optimization updates only $W_{B},W_{A}$ while leaving $W_{0}$ and thus $\theta_{0}$ unchanged.
We denote the resulting model as $f(x;\theta_{0},\phi)$, where task-specific changes are stored compactly in the adapter checkpoint $\phi$.

\textbf{Problem Definition.}
A user requests removal of a small set of interactions $\mathcal{D}_{-r}\subset\mathcal{D}$ from an already-deployed PEFT LLMRec model $f(\cdot;\theta_0,\phi^\star)$ trained on $\mathcal{D}$, where $|\mathcal{D}_{-r}|$ is typically tiny under continual personalization. Let the remaining data be $\mathcal{D}_r=\mathcal{D}\setminus\mathcal{D}_{-r}$. The gold standard for exact unlearning is retrain-on-remain
\[
\phi_r^\star \;=\; \arg\min_{\phi}\; \mathcal{L}_{\mathrm{rec}}(\theta_0,\phi;\mathcal{D}_r),
\]
which is often computationally impractical to run per request. The goal of unlearning is to compute an updated adapter $\tilde{\phi}$ such that $f(\cdot;\theta_0,\tilde{\phi})$ behaves close to $f(\cdot;\theta_0,\phi_r^\star)$ on recommendation outputs, while being (i) \emph{efficient} enough to respond promptly to frequent deletion requests and (ii) \emph{utility-preserving} on the remaining population, measured by minimal degradation of top-$K$ recommendation quality on $\mathcal{D}_r$. Formal mathematical notations are summarized in the Appendix~\ref{Notations}.

\section{Methodology}
\label{sec:model}
In this section, we propose a novel Utility-aware Contrastive AttenuatioN (U-CAN) framework suitable for GenRec tasks. 
We first formalize \textbf{Unlearning under LoRA}, where a frozen backbone is augmented with a trainable adapter and all updates are confined to this adapter. 
Buil ding on this setting, U-CAN comprises three main stages: \textbf{Contrastive Activation} identifies sensitive neurons from activation discrepancies between forgetting and retention sets; \textbf{Utility Significance} uses a structural approximation strategy to assess how parameters contribute to reasoning by combining weight magnitudes with input activation norms; and \textbf{Adaptive Soft Attenuation} performs one-shot soft masking by rescaling weights according to their continuous risk sores. 
The overall framework is illustrated in Figure~\ref{fig:framework}.

\subsection{Unlearning under LoRA}
Unlearning under LoRA applies a low-rank adapter $\phi$ on top of a frozen pre-trained backbone with parameters $\theta_{0}$, yielding the parameterization $f(\cdot;\theta_{0},\phi)$.
All updates and unlearning interventions are confined to $\phi$ while $\theta_{0}$ remains fixed, which provides a consistent interface for subsequent activation analysis and targeted parameter modification. 

A frozen backbone $\theta_{0}$ combined with a trainable adapter $\phi$ defines the LoRA model $f(\cdot;\theta_{0},\phi)$ and its induced conditional distribution $P(\cdot\mid\cdot)$.
A downstream fine-tuning dataset is denoted by $\mathcal{D}$ and is partitioned into a retention set $\mathcal{D}_{r}$ and a forgetting set $\mathcal{D}_{f}$:
\begin{equation}
\mathcal{D}=\mathcal{D}_{r}\cup \mathcal{D}_{f},
\qquad
\mathcal{D}_{r}\cap \mathcal{D}_{f}=\varnothing,
\end{equation}
where $\mathcal{D}_{f}$ corresponds to samples that must be removed. Training on $\mathcal{D}$ yields a deployed adapter $\phi^{\star}$ and the model $f(\cdot;\theta_{0},\phi^{\star})$.

\paragraph{Unlearning Signals}
\label{sec:objective_signals}
Given a frozen backbone with parameters $\theta_{0}$ and a deployed adapter $\phi^{\star}$ trained on $\mathcal{D}=\mathcal{D}_{r}\cup\mathcal{D}_{f}$, LoRA unlearning yields an updated adapter $\tilde{\phi}$ via $\mathcal{U}(\cdot)$:
\begin{equation}
\tilde{\phi}=\mathcal{U}\!\left(\phi^{\star};\,\theta_{0},\mathcal{D}_{r},\mathcal{D}_{f}\right),
\end{equation}
where $\Delta\theta_{0}=0$. To represent inference in a consistent prompting structure, we use a prompt template $\mathcal{T}(\cdot)$ together with a binary token mask $M$ that selects valid positions for token-level computations.

For each layer $l$, let $H^{l}_{t}(x;\theta_{0},\phi)$ denote the activation at token position $t$ produced by
$f(\cdot;\theta_{0},\phi)$ on the templated input $\mathcal{T}(x)$.
We aggregate token-level activations over positions with $M_t=1$ to obtain a layer-wise activation vector:
\begin{equation}
v^{l}(x;\theta_{0},\phi)
\;=\;
\operatorname{Agg}\!\left(\left\{\,H^{l}_{t}(x;\theta_{0},\phi)\;:\; M_t=1\,\right\}\right),
\end{equation}
where $\operatorname{Agg}(\cdot)$ denotes a generic token-wise aggregation operator over positions.
Using the deployed adapter $\phi^{\star}$, we compute the activation summaries on the forgetting and retention subsets as
\begin{align}
v^{l}_{i} &\;\equiv\;\mathbb{E}_{(x)\in\mathcal{D}_{i}}\!\left[v^{l}(x;\theta_{0},\phi^{\star})\right], 
\end{align}
where $i\in\{f,r\}$, which represents the forget and retain sides. These layer-wise signals provide a direct interface for subsequent localization, where candidate parameters are identified by contrasting the activation responses induced by $\mathcal{D}_{f}$ and $\mathcal{D}_{r}$ under the same deployed adapter, and then used to guide localized parameter updates that implement $\mathcal{U}(\cdot)$.

\subsection{Contrastive Activation}
\label{subsection: contrastive}
Sensitive parameter representation identification constitutes a primary challenge in efficient MU for LLMs. High-dimensional spaces induce intense knowledge entanglement, where individual units simultaneously encode private data and general reasoning capabilities. This polysemy thwarts direct localization. To address this, we analyze model activation disparities to identify candidate sensitive regions.

\textbf{Contrastive Activation Feature Extraction.} Compared to raw activation analysis, contrastive activation specifically focuses on isolating neuronal responses that are preferentially associated with $\mathcal{D}_f$. As illustrated in the Figure~\ref{fig:framework}, we initiate the process by collecting token-level activations $H^{l}_{t}(x;\theta_{0},\phi^{\star})$ at each linear layer $l$ via forward propagation on the forgetting set $\mathcal{D}_f$ and retention set $\mathcal{D}_r$, respectively. 

To eliminate input-specific noise and capture stable knowledge representations, we formulate the privacy activation vector $v_{f}^{l}$ and the general activation vector $v_{r}^{l}$ by computing the average response intensity along the token dimension. Crucially, we mask system-prompt tokens so that aggregation depends only on the user interaction sequence, and we instantiate $\operatorname{Agg}(\cdot)$ as the masked average:
\begin{equation}
v_i^{l}
=\mathbb{E}_{x\in \mathcal{D}_{i}}\!\left[
\frac{\sum_{t} M_t\, H^{l}_{t}(x;\theta_{0},\phi^{\star})}{\sum_{t} M_t}
\right],
\end{equation}
where $i\in\{f,r\}$, which represents the forget and retain sides, respectively. $M$ denotes the binary sequence mask ($M_t \in \{0, 1\}$), which filters out system prompts and retains only the tokens corresponding to the user interaction history. 

To characterize the selective response of neurons to privacy features, we formulate a contrastive activation difference metric. For the $j$-th neuron in the $l$-th layer, its preliminary risk score is calculated as follows:
\begin{equation}
r_{gap,j}^{l}=\text{ReLU}(v_{f,j}^{l}-\gamma \cdot v_{r,j}^{l}),
\end{equation}
where $v_{f,j}^{l}$ and $v_{r,j}^{l}$ are the specific activation intensities , and the parameter $\gamma \in \mathbb{R}^{+}$ is defined as the tolerance margin. 
Introducing $\gamma$ yields a dynamic threshold that down-weights the general-activation component, thereby highlighting neurons that respond strongly to privacy signals but weakly to general tasks. 
We further apply $\mathrm{ReLU}$ to retain only positive activation gains and suppress negative-correlation interference. 
The resulting gap scores are min--max normalized as:
\begin{equation}
\tilde{r}_{gap,j}^{l} = \frac{r_{gap,j}^{l} - \min(\mathbf{r}_{gap}^{l})}{\max(\mathbf{r}_{gap}^{l}) - \min(\mathbf{r}_{gap}^{l}) + \epsilon}.
\end{equation}

Subsequently, intra-layer Min-Max normalization is applied to these raw scores, yielding the normalized candidate risk distribution $\tilde{r}_{gap}^{l} \in [0, 1]$.

\subsection{Utility Significance}
To prevent the intensity of privacy sensitivity in a neuron from obscuring the risks associated with the contribution to overall model performance, U-CAN designs independent strategies to evaluate utility significance. Inspired by recent pruning research~\cite{SunLB2024}, U-CAN employs a structural approximation strategy that integrates weight magnitudes with input activation norms to identify the contribution of parameters to reasoning.

Specifically, the importance score $r_{imp,j}^{l}$ for the $j$-th neuron in layer $l$ is calculated as the mean structural sensitivity:
\begin{equation}
r_{imp,j}^{l}
= \frac{1}{d_{out}} \,\lVert W_{\cdot, j}^{l}\rVert_{1}\cdot \lVert X_{\cdot j}^{l}\rVert_{2}
\triangleq \mathbb{E}_{i}\!\left[ \lvert W_{i,j}^{l}\rvert \cdot \sqrt{\sum_{x \in \mathcal{D}_r} \big(H_{j}^{l}(x)\big)^{2} + \epsilon} \right],
\label{eq:r_imp}
\end{equation}
where $W_{\cdot, j}^l$ denotes the column vector corresponding to the $j$-th neuron, and $\|X_{\cdot j}^l\|_2$ represents the $L_2$ norm of input activations accumulated over the retention set $D_r$.
The term $\frac{1}{d_{out}}$ enforces structural column-wise aggregation and defines the $j$-th input dimension as an atomic unit. Since this dimension encodes specific features for user preferences, the utility evaluation corresponds to the feature propagation path.

To align with efficient deployment paradigms, U-CAN incorporates a $4$-bit $NF4$ quantization-aware design.
We define a dynamic dequantization operator $\mathcal{D}(\cdot)$ to transiently recover high-precision weight proxies $\tilde{W}$:
\begin{equation}
\tilde{W}_{\cdot, j}^{l} = \mathcal{D}\!\big(Q(W_{\cdot, j}^{l}), \mathcal{S}_q\big),
\end{equation}
where $Q(\cdot)$ and $\mathcal{S}_q$ denote the quantized weights and quantization state, respectively.
Furthermore, to circumvent GPU memory bottlenecks when processing extensive retention sets, we formulate an IO-aware streaming aggregation protocol.
We partition the retention set $D_r$ into a sequence of mini-batches $D_r = \bigcup_{k=1}^M \mathcal{B}_k$.
The scalar activation norm in Eq.~\ref{eq:r_imp} is then reconstructed via an iterative state update mechanism using a global accumulator vector $\mathbf{S} \in \mathbb{R}^{d_{in}}$:
\begin{equation}
\mathbf{S}^{(k)} \leftarrow \mathbf{S}^{(k-1)} + \sum_{\mathbf{x} \in \mathcal{B}_k} \big(\mathbf{x}^l \odot \mathbf{x}^l\big), \quad
\lVert X_{\cdot j}^{l}\rVert_{2} \approx \sqrt{\mathbf{S}_j^{(K)} + \epsilon},
\end{equation}
where $\odot$ denotes the element-wise product, $\mathbf{S}^{(0)}=\mathbf{0}$, and $\mathbf{x}^l \in \mathbb{R}^{d_{in}}$ represents the activation vector for a single sample.
Finally, the actual importance score is obtained by substituting the dequantized proxy $\tilde{W}$ and the streaming activation norm into Eq.~\ref{eq:r_imp}.

\begin{table*}[htbp]
  \centering
  \caption{Main results (forget vs.\ retain) on ML-100k and Pantry. We report Recall (R), MRR (M), and NDCG (N) at $K\in\{5,10\}$, plus an trade-off score at @10. Best results are in bold. - denotes unavailable results. All unlearning methods operate on LlamaRec and are not compared to LlamaRec itself. ‘Retraining’ denotes full retraining of LlamaRec.}
  \label{tab:accuracy}
  \renewcommand{\arraystretch}{1.2} 
  \setlength{\tabcolsep}{4pt}
  \resizebox{\textwidth}{!}{
  \begin{tabular}{ccc cccccc | cccccc}        
    \toprule
    \multirow{2}{*}{\textbf{Datasets}} & 
    \multirow{2}{*}{\textbf{Models}} & 
    \multirow{2}{*}{\textbf{Trade-off@10} $\uparrow$} &
    \multicolumn{6}{c|}{\textbf{Unlearning Effectiveness} $\downarrow$} & 
    \multicolumn{6}{c}{\textbf{Utility Preservation} $\uparrow$} \\
    \cline{4-15} 
    & & & R@10 & M@10 & N@10 & R@5 & M@5 & N@5 & R@10 & M@10 & N@10 & R@5 & M@5 & N@5 \\ \hline

    \multirow{6}{*}{\centering ML-100k}
      & LlamaRec     & - & 0.2381 & 0.0811 & 0.1176 & 0.1429 & 0.0684 & 0.0868 & 0.1180 & 0.0456 & 0.0623 & 0.0704 & 0.0394 & 0.0471 \\
      \cdashline{4-15}
      & Retraining   & 13.9416 & 0.1999 & 0.0602 & 0.0926 & 0.1200 & 0.0498 & 0.0671 & \textbf{0.1131} & \textbf{0.0422} & \textbf{0.0586} & 0.0626 & 0.0355 & \textbf{0.0421} \\ 
      & GA           & -3.8209 & 0.2303 & 0.0825 & 0.1171 & 0.1569 & 0.0729 & 0.0936 & 0.1030 & 0.0321 & 0.0485 & 0.0622 & 0.0332 & 0.0380 \\
      & NPO          & -4.3017 & 0.2248 & 0.0874 & 0.1198 & 0.1581 & 0.0779 & 0.0977 & 0.1030 & 0.0321 & 0.0485 & 0.0622 & 0.0331 & 0.0379 \\
      & Llama-Eraser & 17.4917 & 0.1714 & 0.0449 & 0.0741 & 0.0838 & 0.0334 & 0.0459 & 0.1032 & 0.0354 & 0.0511 & 0.0639 & 0.0302 & 0.0384 \\
      \rowcolor{gray!20}
      & U-CAN (Ours) & \textbf{29.4548} & \textbf{0.1435} & \textbf{0.0408} & \textbf{0.0639} & \textbf{0.0534} & \textbf{0.0292} & \textbf{0.0352} & 0.1098 & 0.0337 & 0.0514 & \textbf{0.0672} & \textbf{0.0380} & 0.0376 \\ \hline

    \multirow{6}{*}{\centering Pantry}
      & LlamaRec     & - & 0.0416 & 0.0168 & 0.0226 & 0.0287 & 0.0152 & 0.0185 & 0.0464 & 0.0193 & 0.0257 & 0.0332 & 0.0175 & 0.0214 \\
      \cdashline{4-15}
      & Retraining   & 1.5523 & 0.0380 & \textbf{0.0086} & 0.0191 & 0.0228 & \textbf{0.0068} & 0.0148 & 0.0318 & 0.0104 & 0.0153 & 0.0172 & 0.0085 & 0.0107 \\ 
      & GA           & -23.7418 & 0.0416 & 0.0168 & 0.0226 & 0.0283 & 0.0151 & 0.0184 & 0.0416 & 0.0123 & 0.0158 & 0.0131 & 0.0076 & 0.0115 \\
      & NPO          & -23.8512 & 0.0416 & 0.0168 & 0.0226 & 0.0288 & 0.0152 & 0.0186 & 0.0416 & 0.0123 & 0.0157 & 0.0131 & 0.0075 & 0.0114 \\
      & Llama-Eraser & -7.1262 & 0.0347 & 0.0154 & 0.0201 & 0.0275 & 0.0145 & 0.0177 & 0.0406 & 0.0128 & 0.0193 & 0.0252 & 0.0108 & 0.0143 \\
      \rowcolor{gray!20}
      & U-CAN (Ours) & \textbf{14.6441} & \textbf{0.0356} & \textbf{0.0131} & \textbf{0.0184} & \textbf{0.0220} & 0.0113 & \textbf{0.0139} & \textbf{0.0469} & \textbf{0.0177} & \textbf{0.0245} & \textbf{0.0305} & \textbf{0.0154} & \textbf{0.0192} \\ 
    \bottomrule
  \end{tabular}}
\end{table*}
To implement balanced risk hedging, we construct a utility-aware calibration mechanism that recalibrates the preliminary risk scores derived in Section~\ref{subsection: contrastive}. 
For candidate neurons exhibiting high activation discrepancy, their final forgetting weight is adjusted to reconcile privacy erasure with utility preservation. We define the refined risk score $R_{dim}$ through a weighted subtraction process:
\begin{equation}
R_{\text{dim}, j}^{l} = \mathcal{Z}\left( \text{ReLU}\left[ \lambda \cdot \mathcal{N}(\tilde{r}_{\text{gap}, j}^{(l)}) - (1-\lambda) \cdot \mathcal{N}(\tilde{r}_{\text{imp}, j}^{(l)}) \right] \right),
\end{equation}
where $\mathcal{N}(\cdot)$ denotes the Min-Max normalization operator ensuring metrics map to $[0, 1]$, and $\text{ReLU}(x) = \max(0, x)$ eliminates negative risk values derived from high-utility neurons. $\mathcal{Z}(\cdot)$ represents the final re-normalization, and $\lambda$ acts as a balancing coefficient. 

\subsection{Adaptive Soft Attenuation}
Adaptive soft attenuation applies continuous, dimension-wise scaling to suppress privacy-sensitive components without the irreversible connection removal induced by hard pruning.
Instead of binary masks, the intervention is driven by risk scores $R$ through a decay function parameterized by $\alpha$ and $\beta$, which attenuates parameters smoothly as a function of risk magnitude.

Given local risk scores $R_{dim,j}^{l}$, we first select candidate dimensions by thresholding. To account for heterogeneous score scales across layers, we normalize risk scores independently within each layer to $[0,1]$ and define the intervention set
\begin{equation}
\Omega \;=\; \{(l,j)\mid R_{dim,j}^{l}>\tau_{risk}\},
\end{equation}
where $\tau_{risk}$ is the sensitivity threshold.
For $(l,j)\in\Omega$, U-CAN replaces binary gating with score-proportional decay by assigning a retention factor
\begin{equation}
    \alpha_{j}^{l}
    \;=\;
    \alpha_{\text{max}}
    \cdot
    \left(
    1-\frac{R_{dim,j}^{l}-\tau_{risk}}{1-\tau_{risk}+\epsilon}
    \right)^{\beta},
\end{equation}
where $\alpha_{\text{max}}$ bounds the maximum retained magnitude and $\beta$ controls the decay curvature.
This mapping yields a non-linear suppression profile: larger risk scores induce stronger attenuation.

We apply the retention factors as an in-place parameter transformation, using column-wise scaling for each weight dimension. For the weight matrix \( W^l \in \mathbb{R}^{d_{\text{out}} \times d_{\text{in}}} \) at layer \( l \), the retention factor \( \alpha^l \in \mathbb{R}^{d_{\text{in}}} \) is applied to each column \( j \) as follows:
\begin{equation}
W^l_{j} = W^l_{j} \cdot \alpha_j^l,
\end{equation}
where \( W^l_{j} \) refers to the \( j \)-th column of the weight matrix, and \( \alpha_j^l \) is the scaling factor for the \( j \)-th input dimension. This operation is a one-shot transformation and does not require further optimization or backpropagation, making it architecture-agnostic and applicable to PEFT modules. This operation confines the intervention to the adapter parameters, keeping the pretrained backbone unchanged. To minimize inference-time overhead, we fuse the attenuation factors directly into the parameters. as shown by:
\begin{equation}
W^l_{\text{final}} = W^l \odot \left( \alpha^l \right),
\end{equation}
where $\odot$ denotes element-wise multiplication with $\alpha^l$ broadcast along input columns, thereby folding the attenuation factors into $W^l$ to yield the final modified weights. Detailed pseudocodes and theoretical analysis are shown in Appendix~\ref{pseudocode} and \ref{app:theory_ucan}.

\section{Experiment}
In our experiments, we aim to answer the following research questions:
\textbf{RQ1:} Compared with state-of-the-art unlearning methods, can \texttt{U-CAN} achieve more precise unlearning while preserving GenRec performance?
\textbf{RQ2:} What is the contribution of each component in \texttt{U-CAN}?
\textbf{RQ3:} How does \texttt{U-CAN} compare with state-of-the-art unlearning methods in terms of unlearning efficiency?
\textbf{RQ4:} How do different hyperparameters affect the performance of \texttt{U-CAN}?
\subsection{Experiment Setup}
\subsubsection{Datasets.}
ML-100K and Pantry are used as benchmarks from the movie and e-commerce domains.
ML-100K contains 100K user-item interactions, while Pantry is an Amazon Reviews subset on groceries and household supplies with 32{,}992 products.
Following standard sequential recommendation preprocessing~\cite{YueZK2022}, interactions are sorted chronologically, a 5-core filter is applied to retain users and items with at least five interactions, and items without titles are removed.
Details of datasets are in Appendix~\ref{datasets}.

\subsubsection{Benchmarks.}
U-CAN is compared with 1) the deployed backbone LlamaRec~\cite{YueRM2023}, a two-stage sequential recommender comprising a lightweight retriever and a prompt-based LLM reranker; and 2) representative unlearning methods, including Retraining (retraining the backbone on $\mathcal{D}_{r}$ from scratch), GA~\cite{CSC2022}, NPO~\cite{ZhLB2024}, and LLM-Eraser~\cite{ZhangZZ2025}.
The detailed settings of each baseline are provided in Appendix~\ref{benchmarks}.

\subsubsection{Evaluate Metric.}
Our objective is to achieve precise and efficient unlearning for GenRec while preserving recommendation performance. Therefore, we evaluate unlearning outcomes using seven metrics: 
KL divergence~\cite{KuLA1951}, prediction shift~\cite{NeSR2021}, and Perplexity~\cite{BeVD2003neural} (PPL) characterize distributional and behavioral changes induced by unlearning, while @10 ranking metrics (Recall~\cite{MaD2008}, MRR~\cite{VoM1999}, and NDCG~\cite{JKK2002}) capture recommendation-quality variations on both forgetting and retention sides. Trade-off@10 is additionally defined to compactly summarize the forgetting--utility balance implied by the @10 ranking metrics.
In addition, execution time and model throughput quantify the operational efficiency of the unlearning process.
Full metric definitions and implementation details are deferred to the Appendix~\ref{appendix: evaluation}.

\subsubsection{Parameter Setting,}
U-CAN performs unlearning by updating the LoRA modules while keeping the pretrained backbone fixed.
In the \textit{Utility Significance} and \textit{Adaptive Soft Attenuation} stages, the tolerance margin $\gamma$, risk-fusion coefficient $\lambda$, sensitivity threshold $\tau_{\text{risk}}$, maximum retained magnitude $\alpha_{\max}$, and decay curvature $\beta$ are set as follows: $\gamma = 0.5$, $\lambda = 0.3$, $\tau_{\text{risk}} = 0.2$, $\alpha_{\max} = 0.1$, and $\beta = 2.0$.
All experiments are conducted on a cluster equipped with Intel Xeon CPUs and a single NVIDIA V100 GPU.

\subsection{RQ1: Main Results}
The main results are reported in Table~\ref{tab:accuracy}. We evaluate each method along two axes: forgetting effectiveness on the forget set, where lower values indicate stronger erasure, and utility preservation on the retain set, where higher values indicate better recommendation quality. We also report Trade-off@10, a composite metric that rewards strong forgetting with minimal utility degradation.

(1) Retraining is a strong but expensive reference, since it updates the full backbone on the retention set, making deployment in GenRec difficult to scale. In contrast, U-CAN operates in a single pass on adapters without any retraining, yet still matches retraining on overall forgetting–utility balance and surpasses it on several key metrics across both datasets. This indicates that targeted soft attenuation can recover much of the benefit of retraining while avoiding its substantial computational cost.

(2) GA and NPO yield negative Trade-off@10, with marked retain-side utility drops and dataset-dependent forget-set gains, indicating that loss-level gradient updates in an entangled representation space perturb shared parameters beyond privacy-critical regions and distort general recommendation patterns. By contrast, U-CAN derives localized intervention signals from contrastive activations between $\mathcal{D}_f$ and $\mathcal{D}_r$ and applies soft attenuation on LoRA adapters, achieving stronger and more stable forgetting with substantially smaller retain-side degradation.

(3) Llama-Eraser achieves stronger forgetting than gradient-based baselines, but its trade-off scores vary more across datasets, indicating higher utility cost in some settings. This aligns with hard-masking interventions, where removing a coarse subset of parameters also disrupts representations shared with retain behavior and amplifies utility degradation. In contrast, U-CAN combines utility-aware calibration with continuous soft attenuation to more selectively suppress forget-related activations while limiting retain-set drops, and thus yields the most stable forgetting–utility balance in our evaluation.

\subsection{RQ1: Privacy Effectiveness Studies}
\begin{table}[htbp]
    \centering
    \caption{Unlearning results on ML-100k and Pantry. Higher indicates stronger forgetting effect.}
    \label{tab:privacy_effectiveness}
    \renewcommand{\arraystretch}{1.2} 
    \setlength{\tabcolsep}{2pt}
    
    \begin{tabular}{l cccccc}
        \toprule
        \multirow{2}{*}{\textbf{Models}} & \multicolumn{2}{c}{\textbf{KL Divergence} $\uparrow$} & \multicolumn{2}{c}{\textbf{Pred. Shift (\%)} $\uparrow$} & \multicolumn{2}{c}{\textbf{PPL} $\uparrow$} \\
        \cmidrule(lr){2-3} \cmidrule(lr){4-5} \cmidrule(lr){6-7}
        & ML & Pantry & ML & Pantry & ML & Pantry \\
        \midrule
        GA         & 0.00 & 0.00 & 0.10 & 0.56 & 18.91 & 17.74 \\
        NPO        & 0.00 & 0.00 & 0.38 & 0.95 & 18.90 & 17.77 \\
        LLM-erase  & 0.11          & 0.14          & 6.28          & 18.76         & 21.86 & 18.71 \\
        \hline
        \rowcolor{gray!15}
        U-CAN       & \textbf{0.13}          & \textbf{0.41}          & \textbf{6.54}          & \textbf{25.41}         & \textbf{23.83} & \textbf{69.67} \\
        \bottomrule
    \end{tabular}
\end{table}
To distinguish substantive erasure from superficial suppression, Table~\ref{tab:privacy_effectiveness} reports three forgetting-set signals—KL divergence, Prediction Shift, and PPL—that capture output-distribution change and uncertainty. Across both ML-100k and Pantry, U-CAN consistently induces the largest distributional deviation and uncertainty increase on forgotten data, with especially strong effects on Pantry.

(1) GA and NPO show near-zero KL divergence and only marginal prediction shift on both datasets, with PPL almost unchanged, indicating that gradient-based objective updates barely alter the forgetting-set distribution.
Given that machine unlearning aims for the post-unlearning distribution to match that of training without the forget data~\cite{BourtouleCC2021}, this distributional inertia suggests incomplete removal.
In contrast, U-CAN leverages contrastive activations between forget and retain data to localize forget-specific responses and apply targeted intervention, yielding consistently larger KL divergence and prediction shift; this also accords with the security view that memorization is associated with abnormally low PPL on specific sequences~\cite{CarliniTW2021}, as U-CAN drives PPL upward on the forgetting set, reflecting reduced confidence in the removed content.

(2) LLM-erase raises KL divergence and prediction shift over GA and NPO, indicating that selective pruning can change forget-set outputs, yet PPL on Pantry remains close to gradient baselines, suggesting residual high-confidence likelihood on forgotten sequences. In contrast, U-CAN yields the largest prediction shift and a sharp PPL surge on Pantry (69.67), implying a substantial likelihood collapse on the forget set rather than a mild redistribution. We also evaluate basic plot understanding, narrative generation, and recommendation quality retention after applying different unlearning methods, as detailed in Appendix~\ref{app: case}.

\subsection{RQ2: Ablation Studies}
\begin{table*}[t]
    \centering
    \caption{Ablation of U-CAN. Each variant removes one component: “w/o F" without the utility-significance estimation used for risk calibration; "w/o C" without contrastive screening and computes risk scores from forgetting-set activations only; "w/o H" without adaptive soft attenuation with a hard intervention via binary weight zero.}
    \label{tab:ablation}
    \renewcommand{\arraystretch}{1.1} 
    \setlength{\tabcolsep}{4pt}

    \resizebox{\textwidth}{!}{
    \begin{tabular}{l c c c c c c c c c c c c}
        \toprule
        \multirow{3}{*}{\textbf{Variant}} 
        & \multicolumn{6}{c}{\textbf{Privacy Disruption}} 
        & \multicolumn{6}{c}{\textbf{Unlearning / Utility}} \\
        \cmidrule(lr){2-7} \cmidrule(lr){8-13}
        & \multicolumn{2}{c}{\textbf{KL Divergence}$\uparrow$}
        & \multicolumn{2}{c}{\textbf{Pred. Shift (\%)}$\uparrow$}
        & \multicolumn{2}{c}{\textbf{PPL}$\uparrow$}
        & \multicolumn{3}{c}{\textbf{Unlearning Effectiveness$\downarrow$}}
        & \multicolumn{3}{c}{\textbf{Utility Preservation$\uparrow$}} \\
        \cmidrule(lr){2-3}\cmidrule(lr){4-5}\cmidrule(lr){6-7}
        \cmidrule(lr){8-10}\cmidrule(lr){11-13}
        & ML & Pantry & ML & Pantry & ML & Pantry
        & R@10 & M@10 & N@10 & R@10 & M@10 & N@10 \\
        \midrule
        w/o F & 0.06 & 0.16 & 4.06 & 18.10 & 22.23 & 19.97 & 0.2070 & 0.0714 & 0.1031 & \textbf{0.1197} & \textbf{0.0468} & \textbf{0.0636} \\
        w/o C & 0.01 & 0.17 & 1.86 & 17.69 & 19.73 & 18.02 & 0.1502 & 0.0418 & 0.0661 & 0.0770 & 0.0252 & 0.0374 \\
        w/o H & 0.08 & 0.19 & 4.96 & 18.70 & 20.81 & 23.73 & 0.1536 & 0.0422 & 0.0671 & 0.0820 & 0.0254 & 0.0386 \\
        \hline
        \rowcolor{gray!15}
        U-CAN & \textbf{0.13} & \textbf{0.42} & \textbf{6.54} & \textbf{25.41} & \textbf{23.83} & \textbf{69.67} & \textbf{0.1435} & \textbf{0.0408} & \textbf{0.0639} & 0.1098 & 0.0337 & 0.0515 \\
        \bottomrule
    \end{tabular}}
\end{table*}
Table~\ref{tab:ablation} isolates the effect of each U-CAN component. The full configuration induces the strongest distributional shift on the forget side while maintaining the best effectiveness–utility trade-off, suggesting that contrastive localization, utility-aware screening, and soft attenuation are complementary rather than interchangeable.

(1) On the left, full U-CAN attains the largest KL divergence, prediction shift, and PPL on ML-100k and Pantry, with a particularly large PPL increase on Pantry. “w/o C” yields the weakest KL and prediction shift and consistently lower PPL, indicating that raw activation statistics alone provide an insufficient privacy-localization signal. “w/o F” likewise reduces KL, prediction shift, and PPL relative to the full model, consistent with poorer separation between risk-relevant and utility-critical dimensions. “w/o H” markedly suppresses the PPL rise on Pantry, showing that continuous, risk-weighted suppression, rather than mild probability redistribution, drives the high-uncertainty regime on forgotten data.

(2) Table~\ref{tab:ablation} right shows that full U-CAN attains the lowest forget-side Recall@10, MRR@10, and NDCG@10, while preserving utility better than “w/o C“ and “w/o H“.
Both “w/o C“ and “w/o H“ incur a pronounced utility degradation relative to the full model, supporting that contrastive localization and soft attenuation are critical for reducing collateral damage during unlearning. By comparison, “w/o F“ preserves the highest utility but shows the weakest unlearning effectiveness, indicating that disabling utility-significance estimation for risk calibration weakens forgetting.

\subsection{RQ3: Efficiency Studies}
\begin{figure}[htbp]
  \centering
  \includegraphics[width=1.0\linewidth]{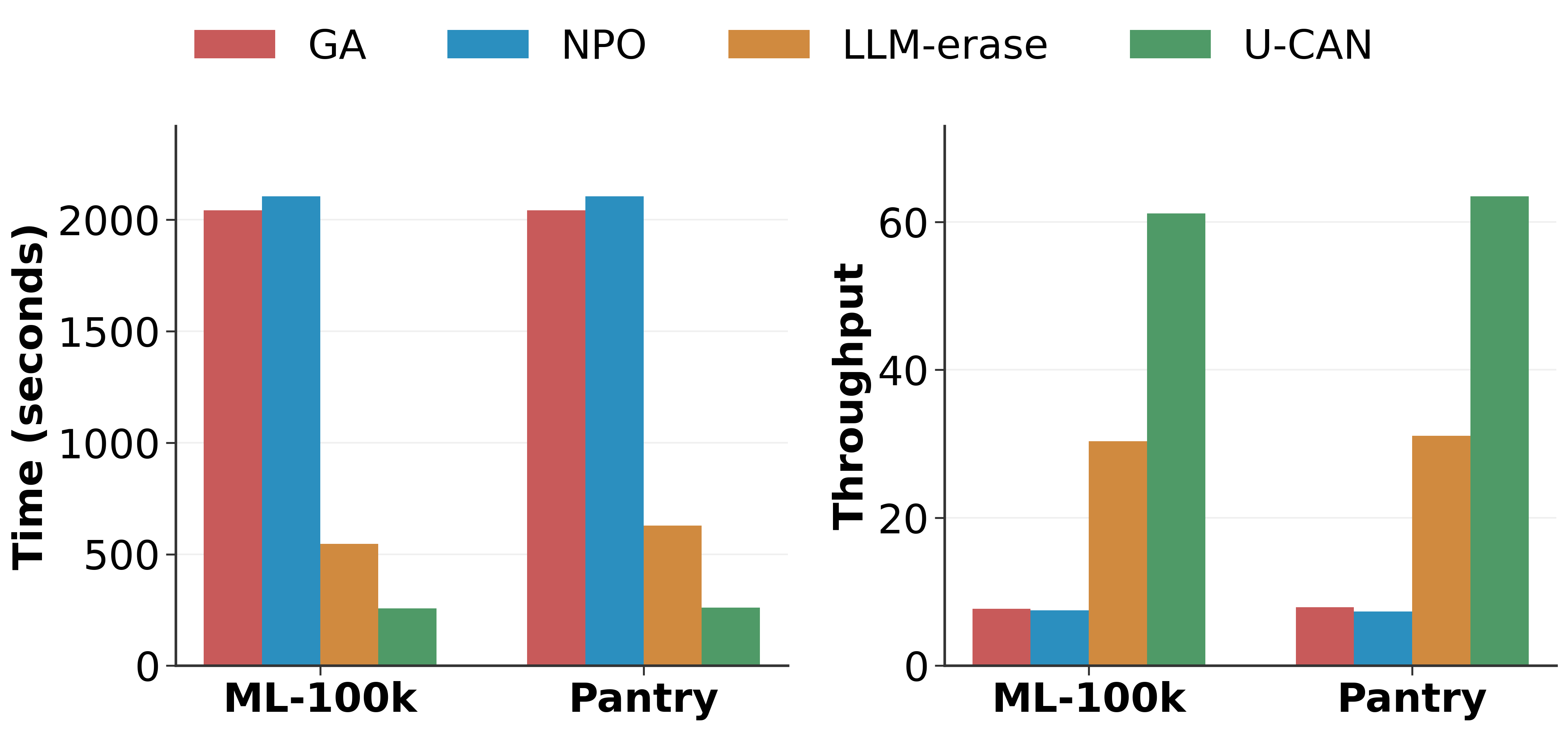}
  \Description{Execution time (left) and throughput (right) for unlearning methods on ML-100k and Pantry.}
  \caption{Execution time (left) and throughput (right) for unlearning methods on ML-100k and Pantry.}
  \label{fig:time}
\end{figure}
Figure~\ref{fig:time} reports execution time and throughput of unlearning baselines and U-CAN on ML-100k and Pantry, characterizing computational efficiency and scalability trade-offs.

(1) Runtime advantage. GA and NPO consistently incur the largest latency on both datasets with near-identical curves, indicating a structural efficiency ceiling: objective-level unlearning requires repeated backward passes and parameter updates under frequent forget requests. In contrast, U-CAN maintains a clear runtime lead by replacing global gradient optimization with forward-only localization, extracting contrastive activation gaps between $\mathcal{D}_f$ and $\mathcal{D}_r$ and performing utility-aware screening in activation space to sidestep the backpropagation bottleneck.

(2) Throughput advantage.
The throughput results mirror the runtime trend. GA and NPO remain throughput-limited, consistent with their reliance on gradient computation over multiple optimization steps. LLM-erase is more efficient than gradient-based baselines, yet throughput still lags behind U-CAN and the runtime gap further widens on Pantry, a pattern consistent with the additional optimization overhead of selective localization compared with contrastive activation-based localization in U-CAN.
U-CAN sustains the highest throughput on both datasets by applying risk-driven, dimension-wise soft attenuation instead of binary masks that disrupt shared representations and require costly recovery.

\subsection{RQ4: Hyper-Parameter Sensitivity Studies}
\begin{figure}[htbp]
  \centering
  \includegraphics[width=1.0\linewidth]{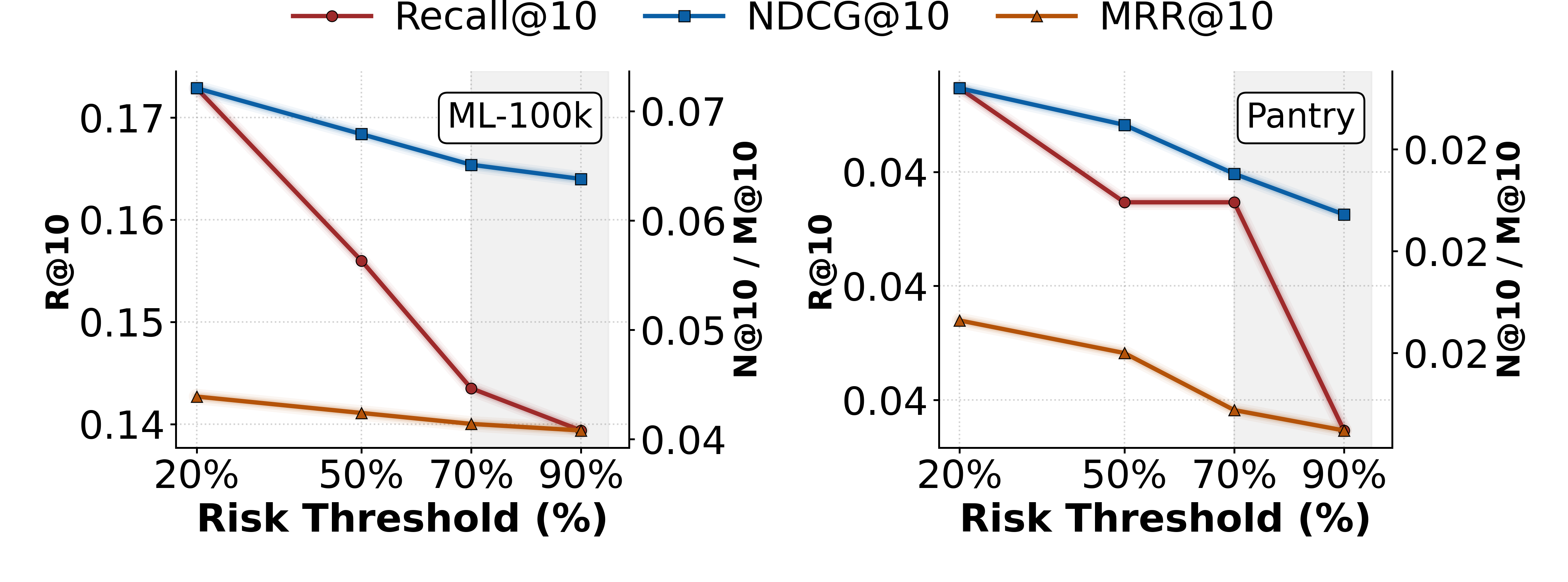}
  \Description{Impact of Risk Threshold on Unlearning Effectiveness.}
  \caption{Impact of Risk Threshold on Unlearning Effectivenes.}
  \vspace{-10pt}
  \label{fig:parameter}
\end{figure}
Figure~\ref{fig:parameter} studies sensitivity to the risk-threshold hyperparameter that controls how aggressively neurons are selected by the dual screening module. Across ML-100k and Pantry, increasing the threshold consistently strengthens forgetting, indicating that U-CAN provides a stable and interpretable knob to trade intervention intensity against residual forget-set performance.

On ML-100k, all three forget-set retrieval metrics decrease as the threshold increases, indicating progressively stronger suppression of forget-associated retrieval behavior under stricter selection. The drop is most pronounced for Recall@10, while MRR@10 and NDCG@10 follow the same direction with smoother declines, suggesting that raising the threshold primarily reduces coarse hit-rate first and then further weakens ranking quality among the remaining hits. 

A similar monotone decline appears on Pantry: increasing $\tau_{risk}$ consistently lowers Recall@10, MRR@10, and NDCG@10, and the curves compress at the strictest setting, suggesting diminishing returns once most high-risk units are already covered. The consistent trends across metrics and datasets indicate that $\tau_{risk}$ provides a stable handle on the fraction of risk-scored neurons selected by dual screening, and that soft attenuation converts broader neuron coverage into deeper forgetting without noticeable oscillation or dataset-specific instability. We further analyze the sensitivity of the fusion weight $\lambda$ on ML-100k, with detailed results reported in Appendix~\ref{appendix: sensitivity parameter}.

\section{Conclusion}
In this paper, we introduce Utility-aware Contrastive Attenuation (U-CAN), an unlearning framework for LLM-based generative recommendation that addresses the tension between precise privacy removal and preserving general capabilities.
U-CAN operates on LoRA adapters and uses a dual screening mechanism: contrastive activation gaps between forget and retain interactions localize privacy-specific neurons, while a utility module combines adapter weight columns with retain-set activations to score the contribution of each hidden dimension to reasoning and recommendation quality. These privacy and utility signals parameterize a differentiable decay function, which an adaptive soft attenuation scheme applies to selectively reduce high-risk dimensions in a single pass, avoiding the structural damage and performance collapse while maintaining the connectivity of shared reasoning pathways.
Experiments on two publicly available datasets assess U-CAN with seven evaluation metrics and indicate strong privacy forgetting, utility retention, and computational efficiency. Appendix~\ref{Limitation} further analyzes the limitations of U-CAN.



\bibliographystyle{ACM-Reference-Format}
\bibliography{sample-base}

@article{LiuYJ2025,
  title={{Rethinking Machine Unlearning for Large Language Models}},
  author={Liu, Sijia and Yao, Yuanshun and Jia, Jinghan and Casper, Stephen and
          Baracaldo, Nathalie and Hase, Peter and Yao, Yuguang and Liu, Chris Yuhao and
          Xu, Xiaojun and Li, Hang and Varshney, Kush R. and Bansal, Mohit and
          Koyejo, Sanmi and Liu, Yang},
  journal={Nature Machine Intelligence},
  volume={7},
  pages={181-194},
  year={2025}
}

@inproceedings{ZhangZZ2025,
  author={Zhang, Shengming and Zhang, Le and Zhou, Jingbo and Zheng, Zhi and Xiong, Hui},
  title={{LLM-Eraser: Optimizing Large Language Model Unlearning through Selective Pruning}},
  booktitle={Proceedings of the 31st ACM SIGKDD International Conference on Knowledge Discovery and Data Mining},
  pages={1960-1971},
  year={2025},
  address = {New York, NY, USA},
  publisher={Association for Computing Machinery},
}

@inproceedings{CaoY2015,
  author={Cao, Yinzhi and Yang, Junfeng},
  title={{Towards Making Systems Forget with Machine Unlearning}},
  booktitle={Proceedings of the 2015 IEEE Symposium on Security and Privacy },
  pages={463-480},
  year={2015},
  address = {USA},
  publisher={IEEE},
}

@inproceedings{BourtouleCC2021,
  author={Bourtoule, Lucas and Chandrasekaran, Varun and Choquette-Choo, Christopher A and Jia, Hengrui and Travers, Adelin and Zhang, Baiwu and Lie, David and Papernot, Nicolas},
  title={{Machine Unlearning}},
  booktitle={2021 IEEE Symposium on Security and Privacy },
  pages={141-159},
  year={2021},
  address={Los Alamitos, CA, USA},
  publisher={IEEE},
}

@article{NguyenHR2025,
  title={{A Survey of Machine Unlearning}},
  author={Nguyen, Thanh Tam and Huynh, Thanh Trung and Ren, Zhao and Nguyen, Phi Le and Liew, Alan Wee-Chung and Yin, Hongzhi and Nguyen, Quoc Viet Hung},
  journal={ACM Transactions on Intelligent Systems and Technology},
  year={2025},
  volume={16},
  pages={108:1-108:46},
}

@inproceedings{GinartGV2019,
  author={Ginart, Antonio A. and Guan, Melody Y. and Valiant, Gregory and Zou, James Y.},
  title={{Making AI Forget You: Data Deletion in Machine Learning}},
  booktitle={Advances in Neural Information Processing Systems},
  pages={3518-3531},
  year={2019},
  publisher={Curran Associates, Inc.},
  address={Red Hook, NY, USA},
}

@inproceedings{JangYY2023,
  author={Jang, Joel and Yoon, Dongkeun and Yang, Sohee and Cha, Sungmin and Lee, Moontae and Logeswaran, Lajanugen and Seo, Minjoon},
  title={{Knowledge Unlearning for Mitigating Privacy Risks in Language Models}},
  booktitle={Proceedings of the 61st Annual Meeting of the Association for Computational Linguistics},
  pages={14389-14408},
  year={2023},
  publisher={Association for Computational Linguistics},
  address={Toronto, Canada},
}

@inproceedings{RafailovSM2023,
  author={Rafailov, Rafael and Sharma, Archit and Mitchell, Eric and Ermon, Stefano and Manning, Christopher D. and Finn, Chelsea},
  title={{Direct Preference Optimization: Your Language Model Is Secretly a Reward Model}},
  booktitle={Advances in Neural Information Processing Systems},
  pages={53728-53741},
  year={2023},
  publisher={Curran Associates, Inc.},
  address={Red Hook, NY, USA},
}

@inproceedings{FanLZ2023,
    author={Fan, Chongyu and Liu, Jiancheng and Zhang, Yihua and Wong, Eric and Wei, Dennis and Liu, Sijia},
    title={{SalUn: Empowering Machine Unlearning via Gradient-based Weight Saliency in Both Image Classification and Generation}},
    booktitle={The Twelfth International Conference on Learning Representations},
    pages={},
    year={2024},
    address={Vienna, Austria},
    publisher={Proceedings of the International Conference on Learning Representations},
}

@inproceedings{LuWH2022,
  author={Lu, Ximing and Welleck, Sean and Hessel, Jack and Jiang, Liwei and Qin, Lianhui and West, Peter and Ammanabrolu, Prithviraj and Choi, Yejin},
  title={{QUARK: Controllable Text Generation with Reinforced Unlearning}},
  booktitle={Advances in Neural Information Processing Systems},
  pages={27591-27609},
  year={2022},
  publisher={Curran Associates, Inc.},
  address={Red Hook, NY, USA},
}

@inproceedings{WuLX2023,
  author={Wu, Xinwei and Li, Junzhuo and Xu, Minghui and Dong, Weilong and Wu, Shuangzhi and Bian, Chao and Xiong, Deyi},
  title={{DEPN: Detecting and Editing Privacy Neurons in Pretrained Language Models}},
  booktitle={Proceedings of the 2023 Conference on Empirical Methods in Natural Language Processing},
  pages={2875-2886},
  year={2023},
  address={Singapore},
  publisher={Association for Computational Linguistics},
}

@inproceedings{MaFW2023,
  author={Ma, Xinyin and Fang, Gongfan and Wang, Xinchao},
  title={{LLM-Pruner: On the Structural Pruning of Large Language Models}},
  booktitle={Advances in Neural Information Processing Systems},
  pages={21702-21720},
  year={2023},
  publisher={Curran Associates, Inc.},
  address={Red Hook, NY, USA},
}

@inproceedings{ZhongWM2023,
  author={Zhong, Zexuan and Wu, Zhengxuan and Manning, Christopher and Potts, Christopher and Chen, Danqi},
  title={{MQuAKE: Assessing Knowledge Editing in Language Models via Multi-Hop Questions}},
  booktitle={Proceedings of the 2023 Conference on Empirical Methods in Natural Language Processing},
  pages={15686-15702},
  year={2023},
  address={Singapore},
  publisher={Association for Computational Linguistics},
}

@inproceedings{GevaSB2021,
  author={Geva, Mor and Schuster, Roei and Berant, Jonathan and Levy, Omer},
  title={{Transformer Feed-Forward Layers Are Key-Value Memories}},
  booktitle={Proceedings of the 2021 Conference on Empirical Methods in Natural Language Processing},
  pages={5484-5495},
  year={2021},
  address={Online and Punta Cana, Dominican Republic},
  publisher={Association for Computational Linguistics},
}

@inproceedings{SunLB2024,
  author={Sun, Mingjie and Liu, Zhuang and Bair, Anna and Kolter, J. Zico},
  title={{A Simple and Effective Pruning Approach for Large Language Models}},
  booktitle={The Twelfth International Conference on Learning Representations},
  pages={},
  year={2024},
  address={Vienna, Austria},
  publisher={Proceedings of the International Conference on Learning Representations},
}

@inproceedings{YueRM2023,
  title={{LlamaRec: Two-Stage Recommendation Using Large Language Models for Ranking}},
  author={Yue, Zhenrui and Rabhi, Sara and Moreira, Gabriel de Souza Pereira and Wang, Dong and Oldridge, Even},
  booktitle={Proceedings of the 1st Workshop on Generative Recommendation at CIKM '23},
  pages={},
  year={2023},
  publisher={Association for Computing Machinery},
  address={Birmingham, United Kingdom}
}

@inproceedings{YueZK2022,
  author={Yue, Zhenrui and Zeng, Huimin and Kou, Ziyi and Shang, Lanyu and Wang, Dong},
  title={{Defending Substitution-Based Profile Pollution Attacks on Sequential Recommenders}},
  booktitle={Proceedings of the 16th ACM Conference on Recommender Systems},
  pages={59-70},
  year={2022},
  publisher={Association for Computing Machinery},
  address={New York, NY, USA},
}

@article{ZhuYW2025,
  title={{Large Language Models for Information Retrieval: A Survey}},
  author={Zhu, Yutao and Yuan, Huaying and Wang, Shuting and Liu, Jiongnan and Liu, Wenhan and Deng, Chenlong and Chen, Haonan and Liu, Zheng and Dou, Zhicheng and Wen, Ji-Rong},
  journal={ACM Transactions on Information Systems},
  volume={44},
  pages={1-54},
  year={2025},
}

@inproceedings{GengLF2022,
  author={Geng, Shijie and Liu, Shuchang and Fu, Zuohui and Ge, Yingqiang and Zhang, Yongfeng},
  title={{Recommendation as Language Processing (RLP): A Unified Pretrain, Personalized Prompt \& Predict Paradigm (P5)}},
  booktitle={Proceedings of the 16th ACM Conference on Recommender Systems},
  pages={299-315},
  year={2022},
  publisher={Association for Computing Machinery},
  address={New York, NY, USA},
}

@inproceedings{CarliniTW2021,
  author={Carlini, Nicholas and Tram{\`e}r, Florian and Wallace, Eric and Jagielski, Matthew and Herbert-Voss, Ariel and Lee, Katherine and Roberts, Adam and Brown, Tom and Song, Dawn and Erlingsson, {\'U}lfar and Oprea, Alina and Raffel, Colin},
  title={{Extracting Training Data from Large Language Models}},
  booktitle={30th USENIX Security Symposium},
  pages={2633-2650},
  year={2021},
  publisher={USENIX Association},
  address={Vancouver, BC, Canada},
 }

@inproceedings{MainiFS2024,
    author={Maini, Pratyush and Feng, Zhili and Schwarzschild, Avi and Lipton, Zachary Chase and Kolter, J. Zico},
    title={{TOFU: A Task of Fictitious Unlearning for LLMs}},
    booktitle={First Conference on Language Modeling},
    pages={},
    year={2024},
    address={Philadelphia, PA, USA},
    publisher={Proceedings of Language Modeling},
}

@inproceedings{DaiDH2022,
  author={Dai, Damai and Dong, Li and Hao, Yaru and Sui, Zhifang and Chang, Baobao and Wei, Furu},
  title={{Knowledge Neurons in Pretrained Transformers}},
  booktitle={Proceedings of the 60th Annual Meeting of the Association for Computational Linguistics},
  pages={8493-8502},
  year={2022},
  address = {Dublin, Ireland},
  publisher = {Association for Computational Linguistics},
}

@inproceedings{NasrCH2023,
    author={Nasr, Milad and Rando, Javier and Carlini, Nicholas and Hayase, Jonathan and Jagielski, Matthew and Cooper, A. Feder and Ippolito, Daphne and Choquette-Choo, Christopher A. and Tram{\`e}r, Florian and Lee, Katherine},
    title={{Scalable Extraction of Training Data from Aligned, Production Language Models}},
    booktitle={The Thirteenth International Conference on Learning Representations},
    pages={},
    year={2025},
    address={Singapore},
    publisher={Proceedings of the International Conference on Learning Representations},
}

@inproceedings{FrankleC2019,
  author={Frankle, Jonathan and Carbin, Michael},
  title={{The Lottery Ticket Hypothesis: Finding Sparse, Trainable Neural Networks}},
  booktitle={The Seventh International Conference on Learning Representations},
  pages={},
  year={2019},
  address={New Orleans, LA, USA},
  publisher={Proceedings of the International Conference on Learning Representation},
}

@inproceedings{WeiWS2022,
  author={Wei, Jason and Wang, Xuezhi and Schuurmans, Dale and Bosma, Maarten and Xia, Fei and Chi, Ed and Le, Quoc V and Zhou, Denny and Chen, Chia-Wei and others},
  title={{Chain-of-Thought Prompting Elicits Reasoning in Large Language Models}},
  booktitle={Advances in Neural Information Processing Systems},
  pages={24824-24837},
  year={2022},
  publisher = {Curran Associates, Inc.},
  address   = {Red Hook, NY, USA}
}

@inproceedings{EldanR2024,
  author={Eldan, Ronen and Russinovich, Mark},
  title={{Who's Harry Potter? Approximate Unlearning for LLMs}},
  booktitle={The Twelfth International Conference on Learning Representations},
  pages={},
  year={2024},
  address={Vienna, Austria},
  publisher={Proceedings of the International Conference on Learning Representations},
}

@article{LiCZ2024,
  title={{Making Recommender Systems Forget: Learning and Unlearning for Erasable Recommendation}},
  author={Li, Yuyuan and Chen, Chaochao and Zheng, Xiaolin and Liu, Junlin and Wang, Jun},
  journal={Knowledge-Based Systems},
  year={2024},
  volume={283},
  pages={111124},
}

@article{ZhangHB2024,
  title={{Recommendation Unlearning via Influence Function}},
  author={Zhang, Yang and Hu, Zhiyu and Bai, Yimeng and Wu, Jiancan and Wang, Qifan and Feng, Fuli},
  journal={ACM Transactions on Recommender Systems},
  year={2024},
  volume={3},
  pages={1-23},
}

@inproceedings{WangZG2025,
  author={Wang, Lingzhi and Zeng, Xingshan and Guo, Jinsong and Wong, Kam-Fai and Gottlob, Georg},
  title={{Selective Forgetting: Advancing Machine Unlearning Techniques and Evaluation in Language Models}},
  booktitle={Proceedings of the AAAI Conference on Artificial Intelligence},
  pages={843-851},
  year={2025},
  publisher={Association for the Advancement of Artificial Intelligence},
  address={Palo Alto, CA, USA},
}

@inproceedings{ANGZW2025,
  author={An, Guojia and Zou, Jie and Wei, Jiwei and Zhang, Chaoning and Sun, Fuming and Yang, Yang},
  title={{Beyond Whole Dialogue Modeling: Contextual Disentanglement for Conversational Recommendation}},
  booktitle={Proceedings of the 48th International ACM SIGIR Conference on Research and Development in Information Retrieval},
  pages={31-41},
  year={2025},
  publisher = {Association for Computing Machinery},
  address   = {New York, NY, USA}
}

@inproceedings{KeDL2024,
  author={Kemper, Sara and Cui, Justin and Dicarlantonio, Kai and Lin, Kathy and Tang, Danjie and Korikov, Anton and Sanner, Scott},
  title={{Retrieval-Augmented Conversational Recommendation With Prompt-Based Semi-Structured Natural Language State Tracking}},
  booktitle={Proceedings of the 47th International ACM SIGIR Conference on Research and Development in Information Retrieval},
  pages={2786-2790},
  year={2024},
  publisher = {Association for Computing Machinery},
  address   = {New York, NY, USA}
}

@inproceedings{WaHZ2025,
  author={Wang, Bo and He, Weiyi and Zeng, Shenglai and Xiang, Zhen and Xing, Yue and Tang, Jiliang and He, Pengfei},
  title={{Unveiling Privacy Risks in LLM Agent Memory}},
  booktitle={Proceedings of the 63rd Annual Meeting of the Association for Computational Linguistics (Volume 1: Long Papers)},
  pages={25241-25260},
  year={2025},
  publisher = {Association for Computational Linguistics},
  address   = {Stroudsburg, PA, USA}
}

@inproceedings{TaoWY2025,
  author={Tao, Yiling and Wang, Shuyi and Yang, Jiaxi and Zuccon, Guido},
  title={{Unlearning for Federated Online Learning to Rank: A Reproducibility Study}},
  booktitle={Proceedings of the 48th International ACM SIGIR Conference on Research and Development in Information Retrieval},
  pages={3377-3386},
  year={2025},
  publisher = {Association for Computing Machinery},
  address   = {New York, NY, USA}
}

@inproceedings{CSC2022,
  author={Chai, Shuwen and Chen, Jinghui},
  title={{One-Shot Neural Backdoor Erasing via Adversarial Weight Masking}},
  booktitle={Advances in Neural Information Processing Systems},
  pages={22285-22299},
  year={2022},
  publisher = {Curran Associates, Inc.},
  address = {Red Hook, NY, USA}
}

@inproceedings{ZhLB2024,
    author={Zhang, Ruiqi and Lin, Licong and Bai, Yu and Mei, Song},
    title={{Negative Preference Optimization: From Catastrophic Collapse to Effective Unlearning}},
    booktitle={First Conference on Language Modeling},
    pages={},
    year={2024},
    publisher = {Proceedings of Language Modeling},
    address={Philadelphia, PA, USA},
}

@article{HuZX2025,
  author={Hu, Zhiyu and Zhang, Yang and Xiao, Minghao and Wang, Wenjie and Feng, Fuli and He, Xiangnan},
  title={Exact and efficient unlearning for large language model-based recommendation},
  journal={IEEE Transactions on Knowledge and Data Engineering},
  year={2025},
  volume={37},
  pages={5866 - 5877}
}

@inproceedings{HuSW2022,
    author={Edward J Hu and Yelong Shen and Phillip Wallis and Zeyuan Allen-Zhu and Yuanzhi Li and Shean Wang and Lu Wang and Weizhu Chen},
    title={Lo{RA}: Low-Rank Adaptation of Large Language Models},
    booktitle={Proceedings of the Tenth International Conference on Learning Representations},
    pages={},
    year={2022},
    publisher = {Proceedings of the International Conference on Learning Representations},
    address   = {Online}
}

@inproceedings{DeHP2023,
    author={Tim Dettmers and Artidoro Pagnoni and Ari Holtzman and Luke Zettlemoyer},
    title={{QL}o{RA}: Efficient Finetuning of Quantized {LLM}s},
    booktitle={Thirty-seventh Conference on Neural Information Processing Systems},
    pages={1-28},
    year={2023},
    publisher = {Neural Information Processing Systems Foundation, Inc.},
    address   = {New Orleans, LA, USA}
}

@inproceedings{LiWY2024,
  author    = {Liu, Shih-Yang and Wang, Chien-Yi and Yin, Hongxu and Molchanov, Pavlo and Wang, Yu-Chiang Frank and Cheng, Kwang-Ting and Chen, Min-Hung},
  title     = {DoRA: weight-decomposed low-rank adaptation},
  booktitle = {Proceedings of the 41st International Conference on Machine Learning},
  pages = {32100--32121},
  year      = {2024},
  series    = {ICML'24},
  volume = {235},
  publisher = {Proceedings of International Conference on Machine Learning},
  address   = {Vienna, Austria}
}

@inproceedings{ChCH2025,
    author={Sungmin Cha and Sungjun Cho and Dasol Hwang and Moontae Lee},
    title={Towards Robust and Parameter-Efficient Knowledge Unlearning for {LLM}s},
    booktitle={The Thirteenth International Conference on Learning Representations},
    pages={},
    year={2025},
    publisher = {Proceedings of the International Conference on Learning Representations},
    address   = {Singapore}
}

@inproceedings{BaZZ2023,
  author={Bao, Keqin and Zhang, Jizhi and Zhang, Yang and Wang, Wenjie and Feng, Fuli and He, Xiangnan},
  title={Tallrec: An effective and efficient tuning framework to align large language model with recommendation},
  booktitle={Proceedings of the 17th ACM conference on recommender systems},
  pages={1007--1014},
  year={2023},
  publisher = {Association for Computing Machinery},
  address   = {Singapore}
}

@inproceedings{GeLF2022,
  author={Geng, Shijie and Liu, Shuchang and Fu, Zuohui and Ge, Yingqiang and Zhang, Yongfeng},
  title={Recommendation as language processing (rlp): A unified pretrain, personalized prompt \& predict paradigm (p5)},
  booktitle={Proceedings of the 16th ACM conference on recommender systems},
  pages={299--315},
  year={2022},
  publisher = {Association for Computing Machinery},
  address   = {Seattle, WA, USA}
}

@inproceedings{GuLY2024,
  author={Guo, Lei and Lu, Ziang and Yu, Junliang and Nguyen, Quoc Viet Hung and Yin, Hongzhi},
  title={Prompt-enhanced federated content representation learning for cross-domain recommendation},
  booktitle={Proceedings of the ACM Web Conference 2024},
  pages={3139--3149},
  year={2024},
  publisher = {Association for Computing Machinery},
  address   = {Singapore}
}

@inproceedings{ZhZW2025,
  title={Towards distribution matching between collaborative and language spaces for generative recommendation},
  author={Zhang, Yi and Zhang, Yiwen and Wang, Yu and Chen, Tong and Yin, Hongzhi},
  booktitle={Proceedings of the 48th International ACM SIGIR Conference on Research and Development in Information Retrieval},
  pages={2006--2016},
  year={2025},
  publisher = {Association for Computing Machinery},
  address   = {Padua, Italy}
}

@inproceedings{YaLW2025,
  title={Earn: Efficient inference acceleration for llm-based generative recommendation by register tokens},
  author={Yang, Chaoqun and Lin, Xinyu and Wang, Wenjie and Li, Yongqi and Sun, Teng and Han, Xianjing and Chua, Tat-Seng},
  booktitle={Proceedings of the 31st ACM SIGKDD Conference on Knowledge Discovery and Data Mining},
  pages={3483--3494},
  year={2025},
  publisher = {Association for Computing Machinery},
  address   = {Toronto, ON, Canada}
}

@inproceedings{LiLY2024,
  title={Llara: Large language-recommendation assistant},
  author={Liao, Jiayi and Li, Sihang and Yang, Zhengyi and Wu, Jiancan and Yuan, Yancheng and Wang, Xiang and He, Xiangnan},
  booktitle={Proceedings of the 47th International ACM SIGIR Conference on Research and Development in Information Retrieval},
  pages={1785--1795},
  year={2024},
  publisher = {Association for Computing Machinery},
  address   = {Washington, DC, USA}
}

@article{pan2024unifying,
  title={Unifying large language models and knowledge graphs: A roadmap},
  author={Pan, Shirui and Luo, Linhao and Wang, Yufei and Chen, Chen and Wang, Jiapu and Wu, Xindong},
  journal={IEEE Transactions on Knowledge and Data Engineering},
  volume={36},
  number={7},
  pages={3580--3599},
  year={2024}
}

@article{naveed2025comprehensive,
  title={A comprehensive overview of large language models},
  author={Naveed, Humza and Khan, Asad Ullah and Qiu, Shi and Saqib, Muhammad and Anwar, Saeed and Usman, Muhammad and Akhtar, Naveed and Barnes, Nick and Mian, Ajmal},
  journal={ACM Transactions on Intelligent Systems and Technology},
  volume={16},
  number={5},
  pages={1--72},
  year={2025}
}

@article{zhao2023survey,
  title={A survey of large language models},
  author={Zhao, Wayne Xin and Zhou, Kun and Li, Junyi and Tang, Tianyi and Wang, Xiaolei and Hou, Yupeng and Min, Yingqian and Zhang, Beichen and Zhang, Junjie and Dong, Zican and others},
  journal={arXiv preprint arXiv:2303.18223},
  volume={1},
  number={2},
  pages={},
  year={2023}
}

@article{KuLA1951,
  title={On information and sufficiency},
  author={Kullback, Solomon and Leibler, Richard A},
  journal={The annals of mathematical statistics},
  volume={22},
  number={1},
  pages={79--86},
  year={1951},
  publisher={JSTOR}
}

@article{BeVD2003neural,
  title={A neural probabilistic language model},
  author={Bengio, Yoshua and Ducharme, R{\'e}jean and Vincent, Pascal and Jauvin, Christian},
  journal={Journal of machine learning research},
  volume={3},
  number={Feb},
  pages={1137--1155},
  year={2003}
}

@inproceedings{NeSR2021,
  title={Descent-to-delete: Gradient-based methods for machine unlearning},
  author={Neel, Seth and Roth, Aaron and Sharifi-Malvajerdi, Saeed},
  booktitle={Algorithmic Learning Theory},
  pages={931--962},
  year={2021},
  publisher = {Proceedings of Machine Learning Research},
  address   = {Cambridge, MA, USA}
}

@book{MaD2008,
  title={Introduction to information retrieval},
  author={Manning, Christopher D},
  year={2008},
  publisher={Cambridge University Press},
  address   = {Cambridge}
}

@inproceedings{VoM1999,
  title={The trec-8 question answering track report.},
  author={Voorhees, Ellen M and others},
  booktitle={Trec},
  volume={99},
  pages={77--82},
  year={1999},
  publisher = {National Institute of Standards and Technology},
  address   = {Gaithersburg, MD, USA}
}

@article{JKK2002,
  title={Cumulated gain-based evaluation of IR techniques},
  author={J{\"a}rvelin, Kalervo and Kek{\"a}l{\"a}inen, Jaana},
  journal={ACM Transactions on Information Systems (TOIS)},
  volume={20},
  number={4},
  pages={422--446},
  year={2002},
  publisher={ACM New York, NY, USA}
}

@inproceedings{0012WYY0Y25,
  author       = {He Zhang and
                  Bang Wu and
                  Xiangwen Yang and
                  Xingliang Yuan and
                  Xiaoning Liu and
                  Xun Yi},
  title        = {Dynamic Graph Unlearning: {A} General and Efficient Post-Processing
                  Method via Gradient Transformation},
  booktitle    = {{WWW}},
  pages        = {931--944},
  publisher    = {{ACM}},
  year         = {2025},
  address   = {Sydney, NSW, Australia}
}

@inproceedings{0012YP24,
  author       = {He Zhang and
                  Xingliang Yuan and
                  Shirui Pan},
  title        = {Unraveling Privacy Risks of Individual Fairness in Graph Neural Networks},
  booktitle    = {{ICDE}},
  pages        = {1712--1725},
  publisher    = {{IEEE}},
  year         = {2024},
  address   = {Utrecht, Netherlands}
}

@article{ZhangWYPTP24,
  author       = {He Zhang and
                  Bang Wu and
                  Xingliang Yuan and
                  Shirui Pan and
                  Hanghang Tong and
                  Jian Pei},
  title        = {Trustworthy Graph Neural Networks: Aspects, Methods, and Trends},
  journal      = {Proc. {IEEE}},
  volume       = {112},
  number       = {2},
  pages        = {97--139},
  year         = {2024}
}

@article{wang2023survey,
  title={A survey on temporal knowledge graph completion: Taxonomy, progress, and prospects},
  author={Wang, Jiapu and Wang, Boyue and Qiu, Meikang and Pan, Shirui and Xiong, Bo and Liu, Heng and Luo, Linhao and Liu, Tengfei and Hu, Yongli and Yin, Baocai and others},
  journal={arXiv preprint arXiv:2308.02457},
  volume={},
  number={},
  pages={},
  year={2023}
}

@article{wang2024large,
  title   = {Large language models-guided dynamic adaptation for temporal knowledge graph reasoning},
  author  = {Wang, Jiapu and Sun, Kai and Luo, Linhao and Wei, Wei and Hu, Yongli and Liew, Alan W and Pan, Shirui and Yin, Baocai},
  journal = {Advances in Neural Information Processing Systems},
  volume  = {37},
  pages   = {8384--8410},
  year    = {2024}
}

@inproceedings{wang2024ime,
  title     = {IME: Integrating Multi-Curvature Shared and Specific Embedding for Temporal Knowledge Graph Completion},
  author    = {Wang, Jiapu and Cui, Zheng and Wang, Boyue and Pan, Shirui and Gao, Junbin and Yin, Baocai and Gao, Wen},
  booktitle = {Proceedings of the ACM Web Conference 2024},
  pages     = {1954--1962},
  year      = {2024},
  publisher = {Association for Computing Machinery},
  address   = {New York, NY, USA}
}

@inproceedings{cheng2025nr4der,
  title={NR4DER: Neural Re-ranking for Diversified Exercise Recommendation},
  author={Cheng, Xinghe and Zhou, Xufang and Fang, Liangda and He, Chaobo and Zhou, Yuyu and Luo, Weiqi and Gong, Zhiguo and Guan, Quanlong},
  booktitle={Proceedings of the 48th International ACM SIGIR Conference on Research and Development in Information Retrieval},
  pages={1738--1747},
  year={2025},
  publisher={Association for Computing Machinery},
  address={New York, NY, USA}
}

@article{cheng2025education,
  title={Education-Oriented Graph Retrieval-Augmented Generation for Learning Path Recommendation},
  author={Cheng, Xinghe and Zhang, Zihan and Wang, Jiapu and Fang, Liangda and He, Chaobo and Guan, Quanlong and Pan, Shirui and Luo, Weiqi},
  journal={arXiv preprint arXiv:2506.22303},
  year={2025},
  eprint={2506.22303},
  archivePrefix={arXiv},
  primaryClass={cs.IR}
}

\appendix
\section*{Appendix}

\section{Model}
\subsection{Mathematical Notations}
\label{Notations}
\begin{table}[ht]
    \centering
    \caption{Summary of Notations}
    \label{tab:notations}
    \begin{tabular}{l|l}
        \toprule
        \textbf{Notation} & \textbf{Description} \\
        \midrule
        $f(\cdot; \theta_0,\phi)$ & LoRA-based GenRec model \\
        $\theta_0$ & Frozen backbone parameters \\
        $\phi / \phi^\star$ & Adapter / Deployed adapter parameters \\
        $P(\cdot)$ & Induced probability distribution \\
        $\mathcal{T}(\cdot)$ & Prompt template \\
        $x / y$ & Textual context / Target item \\
        $\mathcal{D}, \mathcal{D}_f, \mathcal{D}_r$ & Full / Forget / Retain set \\
        $H^l(x;\theta_0,\phi^\star)$ & Layer-$l$ token activations \\
        $v_f^l / v_r^l$ & Privacy / General activation vector at layer $l$ \\
        $M_t$ & Binary mask for system prompts \\
        $\gamma$ & Tolerance margin for filtering \\
        $r_{gap,j}^l$ & Raw activation-gap score of dim $j$ \\
        $\tilde r_{gap,j}^l$ & Normalized gap score of dim $j$ \\
        $W^l_{:,j}$ & Column-$j$ adapter weight at layer $l$ \\
        $\|X_j^l\|_2$ & $L_2$-norm of retain activations on dim $j$ \\
        $r_{imp,j}^l$ & Utility importance score of dim $j$ \\
        $\mathcal{D}(\cdot)$ & Dequantization operator for weights \\
        $R_{dim,j}^l$ & Fused risk score of dim $j$ \\
        $\mathcal{N}(\cdot)$ & Layer-wise Min–Max normalization \\
        $Z(\cdot)$ & Global re-normalization operator \\
        $\lambda$ & Risk–utility fusion coefficient \\
        $\Omega$ & Set of identified high-risk dimensions \\
        $\tau_{risk}$ & Risk sensitivity threshold \\
        $\alpha_j^l$ & Retention factor of dim $j$ at layer $l$ \\
        $\alpha_{\max}$ & Upper bound of retention factor \\
        $\beta$ & Decay curvature parameter \\
        \bottomrule
    \end{tabular}
\end{table}

For clarity and ease of understanding, the mathematical notations used throughout this paper are summarized in Table~\ref{tab:notations}.

\subsection{Pseudocode}
\label{pseudocode}
\begin{algorithm}[ht]
    \caption{Utility-aware Contrastive Attenuation (U-CAN)}
    \label{alg:U-CAN}
    \SetAlgoLined
    \KwIn{Model $\mathcal{M}$ (4-bit NF4), Datasets $\mathcal{D}_f, \mathcal{D}_r$, Params $\gamma, \lambda, \tau, \beta$}
    \KwOut{Unlearned Model $\mathcal{M}^*$}
    
    \tcp{Stage 1: IO-Aware Streaming Statistics}
    Initialize accumulators $\mathbf{v}_f, \mathbf{v}_r, \mathbf{S} \leftarrow \mathbf{0}$\;
    Compute mean activations $\mathbf{v}_f, \mathbf{v}_r$ and utility norm $\mathbf{S} \leftarrow \sum (x \odot x)$ via batched forward passes\;

    \tcp{Stage 2: Layer-wise Risk Evaluation \& Intervention}
    \ForEach{layer $l \in \mathcal{M}$}{
        \tcp{Quantization-Robust Sensitivity Analysis}
        $\tilde{W} \leftarrow \mathcal{D}(Q(W^{(l)}), \mathcal{S}_q)$\;
        $\mathbf{r}_{imp} \leftarrow \frac{1}{d_{out}}\|\tilde{W}\|_1 \odot \sqrt{\mathbf{S}^{(l)}}$\;
        
        \tcp{Risk Fusion and Thresholding}
        $\mathbf{r}_{gap} \leftarrow \text{ReLU}(\mathbf{v}_f^{(l)} - \gamma \mathbf{v}_r^{(l)})$\;
        $\mathbf{R}_{dim} \leftarrow \mathcal{Z}(\text{Fusion}(\mathbf{r}_{gap}, \mathbf{r}_{imp}))$\;
        
        \If{$\mathbf{R}_{dim} > \tau_{risk}$}{
            $\boldsymbol{\alpha} \leftarrow \alpha_{\max} \cdot (1 - \frac{\mathbf{R}_{dim} - \tau_{risk}}{1 - \tau_{risk} + \epsilon})^\beta$ \tcp*{Soft Decay}
            
            \tcp{Architectural Agnostic Update}
            \lIf{is LoRA Adapter}{$W_A \leftarrow W_A \odot \boldsymbol{\alpha}$} 
            \lElse{$W \leftarrow W \odot \boldsymbol{\alpha}$}
            
            $\boldsymbol{\alpha} \leftarrow \mathbb{1}$ \tcp*{Structural Mask Baking}
        }
    }
    \Return $\mathcal{M}$
\end{algorithm}

We illustrate the pseudocode of U-CAN in Algorithm~\ref{alg:U-CAN}.

\subsection{Theoretical Analysis of U-CAN}
\label{app:theory_ucan}
In conducting a theoretical analysis of U-CAN, we examine how its activation- and weight-driven construction functions as an \emph{operational} mechanism for privacy-oriented unlearning, restricting attention to implications that follow directly from the scoring and intervention definitions.

\begin{itemize}
    \item \textbf{Selective activation-gap screening.}
    U-CAN aggregates token activations into masked layer-wise summaries shared by $\mathcal{D}_f$ and $\mathcal{D}_r$, then defines a margin-based gap $\mathrm{ReLU}\big((v_f-v_r)-\gamma\big)$. By nonnegativity, negative forget–retain contrasts are zeroed, so only dimensions with sufficiently positive masked contrast contribute nonzero gap mass and enter downstream risk formation.

    \item \textbf{Risk fusion enables bounded, thresholded selection.}
    U-CAN forms a utility proxy by multiplying the adapter column magnitude with retained-set activation usage, then normalizes it together with the activation-gap proxy and fuses them via weight $\lambda$. Applying $\mathrm{ReLU}$ enforces nonnegativity and re-normalization yields $R_{\mathrm{risk}}$ on a comparable (bounded under min–max normalization) scale; only positive fused values receive nonzero pre-risk mass. The intervention set is then fixed by the rule $R_{\mathrm{risk}}>\tau_{\mathrm{risk}}$, so $\tau_{\mathrm{risk}}$ deterministically controls which dimensions undergo the subsequent parameter rescaling.

    \item \textbf{Adapter-only attenuation via column-wise rescaling.}
    U-CAN freezes the backbone $\theta_0$ and intervenes only on the deployed adapter by assigning each selected dimension a retention factor $\alpha(R_{\mathrm{risk}},\tau_{\mathrm{risk}};\alpha_{\max},\beta)$ and rescaling the corresponding adapter column. By construction, $\alpha\le \alpha_{\max}$ and is monotone non-increasing in $R_{\mathrm{risk}}$ over the intervened range, so higher-risk dimensions are attenuated at least as strongly as lower-risk ones under fixed $(\tau_{\mathrm{risk}},\alpha_{\max},\beta)$. This multiplicative column scaling directly shrinks the adapter’s column-wise contribution for the same inputs, without requiring any gradient-based update.
\end{itemize}

\section{Experiment Setup}
\subsection{Datasets}\label{datasets}
To strictly evaluate the efficacy of privacy erasure and utility preservation, we conduct experiments on two representative benchmarks covering movie and e-commerce domains:
\begin{itemize}
    \item \textbf{ML-100K}\footnote{\url{https://cseweb.ucsd.edu/~jmcauley/datasets/amazon_v2/}}: A widely used benchmark dataset for movie recommendation, containing approximately 100,000 user-item interactions.
    \item \textbf{Pantry}\footnote{\url{https://amazon-reviews-2023.github.io/}}: A subset of the Amazon Review dataset focusing on groceries and household supplies, comprising 32,992 distinct products.
\end{itemize}

Data preprocessing follows standard protocols in sequential recommendation~\cite{YueZK2022}. We organize user interactions chronologically and apply a 5-core filter, recursively retaining only users and items with at least five interactions. Items lacking textual titles are excluded to ensure semantic consistency for the LLM. To simulate a stringent unlearning scenario where users request removal of part of their history, we randomly sample 25\% of user interaction records on each dataset as the forgetting set $\mathcal{D}_f$, and use the remaining 75\% as the retention set $\mathcal{D}_r$.

\subsection{Benchmarks}\label{benchmarks}
We evaluate our framework against a comprehensive set of baselines, including the backbone model and the state-of-the-art unlearning strategies:

\textit{1) Backbone Framework:}
\begin{itemize}
    \item \textbf{LlamaRec}~\cite{YueRM2023}: A two-stage sequential recommender that integrates a lightweight retriever with a prompt-based LLM reranker. It maps logits to candidate probabilities via a verbalizer, ensuring efficient inference. We utilize LlamaRec as the foundation for our privacy pruning research, employing \texttt{Llama-2-7b}\footnote{\url{https://huggingface.co/meta-llama/Llama-2-7b}} as the base generator. Its resource-constrained nature lends significant engineering value to privacy studies in practical deployment scenarios.
\end{itemize}

\textit{2) Unlearning Baselines:}
\begin{itemize}
    \item \textbf{Retraining}: The model is retrained from scratch solely on the retention set ($D_r$). This approach represents the theoretical upper bound for unlearning effectiveness (zero privacy leakage) and serves as the benchmark for utility preservation.
    \item \textbf{GA} (Gradient Ascent)~\cite{CSC2022}: A fundamental unlearning approach that reverses the training objective. It maximizes cross-entropy loss on the forgetting set to shift parameters from the target distribution.
    \item \textbf{NPO} (Negative Preference Optimization)~\cite{ZhLB2024}: Reformulates unlearning as a negative alignment task. It employs a sigmoid-bounded objective to suppress target information while regularizing updates to maintain model stability.
    \item \textbf{LLM-Eraser}~\cite{ZhangZZ2025}: A state-of-the-art framework that localizes undesired memories via neuron scoring and soft masks. It combines selective pruning with contrastive distillation to erase sensitive knowledge without compromising general capabilities.
\end{itemize}

\subsection{Evaluation Indicators}
\label{appendix: evaluation}
To strictly assess the trade-off between privacy erasure and knowledge preservation, we adopt a multi-dimensional evaluation protocol covering recommendation utility, distributional divergence, and generation uncertainty.

\subsubsection{Utility and Performance Metrics}
For the recommendation task, we employ three standard metrics: \textbf{Recall@K}, \textbf{NDCG@K} (Normalized Discounted Cumulative Gain), and \textbf{MRR@K} (Mean Reciprocal Rank). In our experiments, $K$ is set to 10.
\begin{itemize}
    \item \textbf{Recall@K} measures the proportion of relevant items successfully identified within the top-$K$ recommendations.
    \item \textbf{NDCG@K} evaluates the ranking quality by emphasizing hits at higher ranks, reflecting the position bias inherent in user browsing.
    \item \textbf{MRR@K} focuses on the reciprocal rank of the first relevant item, indicating the model's ability to prioritize the ground truth.
\end{itemize}

\textbf{Trade-off@10} is additionally defined to compactly summarize the forgetting--utility balance implied by the @10 ranking metrics.
The specific calculations are as follows:
\begin{equation}
\mathrm{Trade-off@10}=\Delta\%\mathrm{Forget@10}-\Delta\%\mathrm{Retain@10},
\end{equation}
\begin{align}
\Delta\%\mathrm{Forget@10}&=\frac{E_{10}(\theta_o)-E_{10}(\theta_u)}{E_{10}(\theta_o)},
\\
\Delta\%\mathrm{Retain@10}&=\frac{U_{10}(\theta_o)-U_{10}(\theta_u)}{U_{10}(\theta_o)},
\end{align}
where $E_{10}(\theta_o)$ and $U_{10}(\theta_o)$ represent the average @10 scores (Recall, MRR, NDCG) on the forget and retain sets for the original model, while $E_{10}(\theta_u)$ and $U_{10}(\theta_u)$ denote those of the unlearned model.

It is crucial to note that the interpretation of these metrics depends on the target data subset. On the \textbf{Retention Set ($D_r$)}, higher values indicate better preservation of general reasoning capabilities and generalization. Conversely, on the \textbf{Forgetting Set ($D_f$)}, significantly lower values denote effective disruption of the retrieval pathways associated with sensitive information.

\subsubsection{Operational Efficiency Metrics}
To evaluate the scalability of unlearning frameworks in real-world deployment scenarios, we monitor computational overhead through three distinct dimensions:

\textbf{Execution Time.} We record the total wall-clock time required to complete the unlearning process on the specified forgetting set. This metric captures the end-to-end latency, encompassing forward passe and parameter updates.

\textbf{Model Throughput.} Defined as the number of data samples processed per second (samples/sec) during the unlearning phase. Unlike raw execution time, throughput normalizes performance against dataset size, offering a direct measure of the algorithm's processing speed and its capacity to handle high-frequency data deletion requests.

\subsubsection{Unlearning Verification Metrics}
Beyond surface-level performance drops, we introduce three specific indicators to quantify the depth of memory erasure and the distributional shift of the unlearned model $M_{\theta^*}$ compared to the original model $M_{\theta}$.

\textbf{Kullback-Leibler (KL) Divergence.} To measure the distributional distance between the unlearned model and the reference state , we compute the KL Divergence on the forgetting set. 
A higher divergence from the original model when processing sensitive queries suggests the unlearning procedure has successfully altered the model's statistical profile, detaching it from the memorized private data.
\begin{equation}
    D_{KL}(P_{\theta} || P_{\theta^*}) = \sum_{y \in \mathcal{V}} P_{\theta}(y|x) \log \frac{P_{\theta}(y|x)}{P_{\theta^*}(y|x)}
\end{equation}

\textbf{Prediction Shift.} This metric quantifies the tangible behavioral change of the model by calculating the percentage of input samples in $D_f$ where the top-1 generated token or recommended item changes after the unlearning operation. A high Prediction Shift indicates that the model's deterministic preference for the sensitive data has been fundamentally altered.
\begin{equation}
    \text{Shift} = \frac{1}{|D_f|} \sum_{x \in D_f} \mathbb{I}(\arg\max P_{\theta}(y|x) \neq \arg\max P_{\theta^*}(y|x))
\end{equation}

\textbf{Perplexity (PPL).} We utilize Perplexity to evaluate the model's uncertainty regarding the forgotten sequences. Since LLMs tend to exhibit anomalously low perplexity for memorized training data, a sharp increase in PPL on the Forgetting Set implies that the specific parametric memories have been dissolved into high-entropy noise, rendering the information resistant to extraction.
\begin{table}[htbp]
    \centering
    \caption{Hyperparameter Sensitivity Analysis of Fusion Weight $\lambda$ on ML-100k. The optimal configuration is highlighted in bold.}
    \label{tab:lambda_sensitivity}
    \renewcommand{\arraystretch}{1.3} 
    \setlength{\tabcolsep}{1.0pt}
    
    \begin{tabular}{l ccc ccc}
        \toprule
        \multirow{2}{*}{\textbf{Value}} & \multicolumn{3}{c}{\textbf{Unlearning Effectiveness}} & \multicolumn{3}{c}{\textbf{Utility Preservation}} \\
        \cmidrule(lr){2-4} \cmidrule(lr){5-7}
        & R@10 & M@10 & N@10 & R@10 & M@10 & N@10 \\
        \midrule
        $\lambda=0.9$      & 0.2270 & 0.0802 & 0.1145 & 0.1180 & 0.0432 & 0.0606 \\
        $\lambda=0.7$      & 0.2237 & 0.0803 & 0.1140 & 0.1180 & 0.0431 & 0.0605 \\
        $\lambda=0.5$      & 0.1869 & 0.0641 & 0.0925 & 0.1049 & 0.0310 & 0.0480 \\
        \rowcolor{gray!15}
        \textbf{$\lambda=0.3$} & \textbf{0.1435} & \textbf{0.0408} & \textbf{0.0639} & \textbf{0.1098} & \textbf{0.0337} & \textbf{0.0515} \\
        \bottomrule
    \end{tabular}
\end{table}
\begin{table*}[htbp]
  \centering
  \caption{Qualitative comparison of basic narrative understanding and structured narrative generation. 
\textcolor{teal}{\textbf{Teal}} text highlights successful behavior, and red text highlights failure cases.}
  \label{tab:case_study_main}
  \begin{tabular}{L{0.23\textwidth} L{0.23\textwidth} L{0.23\textwidth} L{0.23\textwidth}}
    \toprule
    \textbf{User Instruction} & \textbf{GA} & \textbf{LLM-Erase} & \textbf{U-CAN (Ours)} \\
    \midrule
    \textit{``Explain the plot of the movie Toy Story.''}
    & Returns: \textit{``Toy Story is about a group of toys who come to life when humans are not around. The main character, Woody, is a cowboy doll who is Andy's favorite toy. When Buzz Lightyear, a space ranger action figure, comes to town, he becomes Andy's new favorite toy and Woody feels threatened.''} \newline \textcolor{red}{(Failure: Accurately captures Toy Story’s core premise and inciting conflict but omits the later plot and resolution.)}
    & Returns: \textit{``Woody is jealous that Buzz Lightyear has become his boy's favorite new toy, so he tries to sabotage him until they team up against Sid and Woody learns how special the kid really loves him...I think?''} \newline \textcolor{red}{(Failure: Linguistic degradation.)}
    & Returns: \textit{``Toy Story is a 1995 American computer-animated buddy comedy film produced by Pixar Animation Studios and released by Walt Disney Pictures. It was directed by John Lasseter from a screenplay by Pixar's director team, the Andersen/Lasseter/Gruwell trio, from a story by Lasseter and Pete Docter. The film was Pixar's first feature-length film and was originally released on November 22, 1995. It was the first fully computer-animated feature film.''} \newline \textcolor{teal}{(Success: Maintains high fluency and factual correctness, showing no signs of model collapse.)} \\
    \hline
    \textit{``Explain the plot of the movie Toy Story in five logically ordered sentences, making sure each sentence follows naturally from the previous one.''}
    & Returns: \textit{``The movie Toy Story is about a group of toys that come to life when humans are not around. The main character is Woody, a cowboy doll who is Andy's favorite toy. When Andy gets a new toy, Buzz Lightyear, Woody becomes jealous and tries to get rid of him. However, Woody eventually realizes that he needs Buzz and the two become friends.''} \newline \textcolor{red}{(Failure: Mostly accurate and coherent summary of the main plot, but only in four sentences instead of the requested five.)}
    & Returns: \textit{``The movie Toy Story is about a cowboy doll named Woody who is jealous when his owner Andy gets a new toy called Buzz Lightyear. Woody tries to get rid of Buzz but fails and they become friends. Woody's jealousy comes back when Buzz becomes more popular than him. Woody tries to get rid of Buzz again but fails and they become friends again.''} \newline \textcolor{red}{(Failure: Captures Woody’s jealousy toward Buzz and their eventual friendship in a roughly ordered way, but uses only four sentences.)}
    & Returns: \textit{``The story is about a cowboy doll named Woody who lives with Andy and his family. He has to compete for attention against Buzz Lightyear, another new toy that was given as a gift on Andy's birthday. When they are both accidentally thrown out into the yard by Andy, Woody tries to get back inside before he gets lost or destroyed while Buzz goes off looking for other people to play with him. Eventually, after many adventures together (and some alone), they become friends instead of enemies like at first when it seemed like there could only be one winner between them because no matter what happened during their time apart from each other; whether good things happen or bad ones do too - everything always ends up being okay again eventually!''} \newline \textcolor{teal}{(Success: Captures the key characters, conflict, and overall story arc.)} \\
    \bottomrule
  \end{tabular}
\end{table*}
\begin{table*}[htbp]
  \centering
  \caption{Qualitative comparison on compositional reasoning for audience-specific recommendations.}
  \label{tab:case_study_comp}
  \begin{tabular}{L{0.23\textwidth} L{0.23\textwidth} L{0.23\textwidth} L{0.23\textwidth}}
    \toprule
    \textbf{User Instruction} & \textbf{GA} & \textbf{LLM-Erase} & \textbf{U-CAN (Ours)} \\
    \midrule
    \textit{``List two family-friendly Pixar movies that are suitable for a 10-year-old child. For each movie, write one short sentence explaining why it is appropriate for children.''}
    & Returns: \textit{`` Pixar''} \newline \textcolor{red}{(Failure: Ignoring to satisfy both the request to list two specific movies and to justify why they are appropriate for children.)}
    & Returns: \textit{``[Toy Story] - The first Pixar movie ever made, and still one of the best. It's a great introduction to the world of animation.
[Up] - A heartwarming story about a man who loses his wife but finds love again.
''} \newline \textcolor{red}{(Failure: Selects two suitable Pixar films with plausible reasons, but grounds them in historical and adult-focused themes rather than child-appropriate criteria.)}
    & Returns: \textit{``[Toy Story] - The first Pixar movie was released in 1995 and was the first feature-length computer-animated film. It tells the story of a group of toys who come to life when humans aren't around. The movie has a lot of humor and adventure, and it teaches kids about friendship and teamwork.
[Up] - Released in 2009, Up is another Pixar classic that tells the story of an elderly man who goes on an adventure with a young boy. The movie has a lot of heart and teaches kids about overcoming obstacles and following their dreams.''} \newline \textcolor{teal}{(Success: Correctly lists two family-friendly Pixar movies and provides appropriate child-oriented justifications for each.)} \\
    \bottomrule
  \end{tabular}
\end{table*}

\section{Experiments}
\subsection{RQ4: Sensitivity Analysis of Fusion Parameter}
\label{appendix: sensitivity parameter}
The fusion parameter $\lambda$ acts as the main regulator of our selection mechanism, controlling the relative weight of contrastive activation gaps and utility significance in the fused risk score. To examine how sensitive U-CAN is to this trade-off, we vary $\lambda$ and report the corresponding unlearning and utility metrics on ML-100k in Table~\ref{tab:lambda_sensitivity}.

Table~\ref{tab:lambda_sensitivity} reveals a clear inverse relationship between the fusion weight $\lambda$ and unlearning effectiveness on ML-100k. As $\lambda$ decreases from $0.9$ to $0.3$, the forget-set retrieval metrics (R@10/M@10/N@10) drop monotonically, indicating progressively stronger suppression of privacy-sensitive behaviour. This pattern is consistent with the Polysemy Dilemma. For large $\lambda$, the fused score is dominated by raw activation gaps and thus flags not only privacy-specific units but also polysemous dimensions that are heavily reused by general recommendation tasks. When $\lambda$ is reduced, the utility-aware component has greater influence and down-weights neurons that are important for retain-side performance, filtering out many such false positives. The remaining high-risk set is therefore concentrated on dimensions with strong privacy activation but limited contribution to core utility. Under the optimal setting $\lambda=0.3$, this screening produces the lowest forget-side retrieval scores while keeping utility-preservation metrics close to the baseline, suggesting that U-CAN can intervene more surgically on genuinely privacy-critical parameters without incurring noticeable degradation on the retain set.

\subsection{Qualitative Case Studies}
\label{app: case}
Table~\ref{tab:case_study_main} examines whether unlearning methods preserve basic plot understanding and structured narrative generation after unlearning. 
Table~\ref{tab:case_study_comp} evaluates compositional, audience-specific recommendation for a child user, probing how well each method balances content safety with explanatory recommendation quality.

(1) As shown in Table~\ref{tab:case_study_main}, for both the free form plot explanation and the five sentence structured summary of \textit{Toy Story}, GA and LLM Erase capture the main premise and the jealousy to reconciliation arc. However, their outputs are often short, only loosely ordered, or linguistically degraded, and they frequently fail to satisfy the requested sentence structure or to cover the full story arc. In contrast, U CAN generates more complete and coherent narratives that introduce the key characters, conflict, development, and resolution, and it follows a clearer sentence to sentence progression, suggesting that unlearning does not cause an obvious collapse in basic narrative planning.

(3) As shown in Table~\ref{tab:case_study_comp}, in the audience-specific recommendation task for a 10-year-old child, GA fails to follow the instruction at all, returning only ``Pixar'' without concrete titles or justifications. LLM-Erase lists plausible Pixar films with reasons grounded in historical or adult-oriented themes rather than child-appropriate criteria. U-CAN, however, recommends two family-friendly Pixar movies and explicitly motivates them with child-centric factors such as humor, adventure, friendship, and teamwork, demonstrating better compositional reasoning about both content suitability and explanation structure.

\section{Limitation}
\label{Limitation}
Although U-CAN shows promising results on GenRec with LoRA adapters, it has certain limitations. 
First, our study confines the privacy evaluation to forgetting-set metrics and empirical extraction or inference tests rather than formal guarantees such as differential privacy. 
This leaves open whether the observed benefits carry over to alternative datasets or stricter privacy notions. 
Second, we have not yet systematically assessed U-CAN in other domains, modalities, or more complex reasoning tasks, nor against stronger adaptive adversaries. 
Extending the framework along these axes remains an important direction for future work.

\end{document}